\documentclass[10pt, a4paper]{article}

\usepackage{subfiles}
\usepackage[utf8]{inputenc}
\usepackage[T1]{fontenc}
\usepackage[T1]{fontenc}
\usepackage{lmodern}
\usepackage{amsmath, amssymb, amsfonts, mathtools}
\usepackage{amsmath,amssymb}
\usepackage{bm}
\usepackage{textcomp}
\usepackage{lineno}
\usepackage{multirow}
\usepackage{graphicx}
\usepackage{booktabs}
\usepackage{longtable}
\usepackage{hyperref}
\usepackage{algorithm}
\usepackage{algorithmic}
\usepackage{xcolor}
\usepackage{lineno}
\usepackage{caption}
\usepackage{geometry}
\usepackage[numbers]{natbib}
\usepackage{doi}
\usepackage{setspace}
\usepackage{enumitem}
\usepackage{comment}
\usepackage{tikz}
\newcommand{\reg}{\textsuperscript{\fontsize{3}{3}\selectfont\textregistered}}
\usetikzlibrary{positioning, arrows.meta}
\usetikzlibrary{decorations.pathreplacing, positioning}
\usetikzlibrary{shapes.callouts, shapes.geometric, backgrounds}
\usetikzlibrary{calc, patterns}
\usetikzlibrary{fit}
\usepackage{pgffor}
\usepackage[protrusion=false]{microtype}
\usepackage[affil-it]{authblk}
\usepackage{siunitx}[=v2]
\usepackage{cleveref}
\newcommand{\SelectedLegendLine}[1]{%
  \tikz[baseline=-0.5ex]{\draw[#1] (0,0)--(0.7,0);}%
}

\pgfmathsetmacro{\k}{16.27}
\pgfmathsetmacro{\L}{0.06}

\pgfmathdeclarefunction{square}{1}{%
      \pgfmathparse{#1^2}%
}
\pgfmathdeclarefunction{cube}{1}{%
      \pgfmathparse{#1^3}%
}
\setcitestyle{numbers} 
\newcommand{\Bi}{\mathrm{Bi}}

\newcommand{\pd}[2]{\frac{\partial #1}{\partial #2}}
\newcommand{\pdd}[2]{\frac{\partial^2 #1}{\partial #2^2}}

\newcommand{\Gc}{\Gamma_{c}}

\newcommand{\dO}{d\Omega}
\newcommand{\dG}{d\Gamma}
\newcommand{\btheta}{\boldsymbol{\Theta}}
\newcommand{\blambda}{\boldsymbol{\Lambda}}

\hypersetup{
	colorlinks=true,
	linkcolor=blue,
	citecolor=blue,
	urlcolor=blue
}

\title{\textbf{Generalized PINN-Based Inverse Estimation of Spatially Varying Heat Transfer Coefficients with Transfer Learning}}

\author[1]{Pradipan Maitra \thanks{Email:\texttt{132403001@smail.iitpkd.ac.in}}}
\author[1]{Jariful Hassan \thanks{Email:\texttt{jariful0786@gmail.com}}}
\author[1]{Arijit Hazra \thanks{\,Corresponding author. Email:\texttt{ahazra@iitpkd.ac.in}}}

\affil[1]{\small Department of Mechanical Engineering, Indian Institute of Technology Palakkad}

\date{}

\begin{document}
\sloppy
\maketitle
\tableofcontents


\begin{abstract}
	This study investigates generalized boundary representations and transfer
	learning for physics-informed neural-network (PINN) estimation of spatially
	varying convective heat transfer coefficients (HTCs) in jet impingement
	cooling. Piecewise-linear (PWL) and neural-network boundary estimators are
	inferred from transient interior-temperature measurements and evaluated
	against CFD--CHT reference HTC profiles at Reynolds numbers of \numlist{5000;10000;15000}. A discrete Adjoint-FEM formulation with gradient regularization
	based on the $H^1$-seminorm provides an independent deterministic
	inverse-method benchmark. Transfer learning is investigated under two source--target conditions: transfer between noise levels at the same Reynolds number, and transfer from a controlled Gaussian-profile problem to CFD--CHT-derived target problems at different Reynolds numbers. For the
	representative $\mathrm{Re}=5000$ target with \SI{5}{\percent} measurement noise,
	quadratic, PWL, and neural-network PINNs achieve relative HTC-profile
	errors of \SIlist{5.08;5.31;5.19}{\percent}, respectively, while generalized
	representations reduce the error for the $\mathrm{Re}=15000$ target from \SI{16.76}{\percent} for the quadratic PINN to \SI{14.41}{\percent} and \SI{14.46}{\percent}. Transfer learning
	reduces target-training time by \SIrange{57.86}{81.71}{\percent} for fixed-Reynolds-number
	transfer and by \SIrange{50.59}{84.94}{\percent} for Gaussian-to-CHT transfer. Its effect on
	reconstruction accuracy is target dependent: fixed-profile PWL transfer
	remains within 0.30 percentage points of scratch training, whereas
	Gaussian-to-CHT transfer produces differences of 0.10--1.52 percentage
	points across the investigated targets and representations. These results
	demonstrate that generalized PINN representations can accommodate spatially
	varying HTC profiles without a prescribed quadratic form, while transfer
	learning can substantially reduce target-training cost for repeated inverse
	estimation, with the resulting reconstruction accuracy depending on
	source--target compatibility.
    
\end{abstract}



\section*{Nomenclature}
\label{sec:nomenclature}
\begingroup
\small
\setlength{\LTleft}{0pt}
\setlength{\LTright}{0pt}
\setlength{\LTpre}{0.4em}
\setlength{\LTpost}{0.6em}
\renewcommand{\arraystretch}{1.08}

\addtocounter{table}{-1}
\begin{longtable}{@{}>{\raggedright\arraybackslash}p{0.23\linewidth}>{\raggedright\arraybackslash}p{0.73\linewidth}@{}}
\endfirsthead
\textbf{Nomenclature (continued)} & \textbf{Definition} \\
\endhead
$\mathbf{A}$ & Assembled backward-Euler forward-system matrix \\
$\mathbf{b},\ b_j$ & Boundary-control vector and nodal Biot-number coefficients \\
$\mathbf{B}_{fc},\mathbf{B}_{fr}$ & Top- and right-boundary Robin matrices \\
$\Bi_{fc},\Bi_{fr}$ & Top- and right-boundary Biot numbers\\
$\Bi_{fc}^G$ & Top-boundary Gaussian Biot number profile \\
$\mathbf{c},\ c_r$ & Quadratic boundary-profile coefficient vector and coefficients \\
$c_p$ & Specific heat capacity of the substrate [$\mathrm{J\,kg^{-1}\,K^{-1}}$] \\
$\mathcal{D}_{\Gamma_c}$ & Set of finite-element control degrees of freedom on the top boundary \\
$h,\ h_{fc},\ h_{fr}$ & HTC, spatially varying top HTC, and fixed right HTC [$\mathrm{W\,m^{-2}\,K^{-1}}$] \\
$H,\ W$ & Thickness and radial extent of the reduced plate domain [$\mathrm{m}$] \\
$\mathbf{H}$ & Observation operator mapping the discrete temperature state to sensors \\
$H^1(\Omega),\ V_h$ & First-order Sobolev space and conforming finite-element space \\
$\mathcal{J},\ J_k$ & Adjoint inverse objective and its value at optimization iterate $k$ \\
$\mathcal{J}_{\mathrm{data}},\mathcal{J}_{\mathrm{reg}}$ & Data-misfit functional and boundary-profile regularization penalty \\
$k$ & Thermal conductivity of the substrate [$\mathrm{W\,m^{-1}\,K^{-1}}$] \\
$\mathbf{K},\mathbf{M}$ & Finite-element conduction stiffness and mass matrices \\
$\mathcal{K}_{\mathrm{obs}}$ & Set of observed time-step indices \\
$L_r$ & Plate aspect ratio, $H/W$ \\
$\mathcal{L}$ & Total PINN loss \\
$\mathcal{L}_{\mathrm{data}},\mathcal{L}_{\mathrm{PDE}}$ & Observation-mismatch and governing-equation losses \\
$\mathcal{L}_{\mathrm{IC}},\mathcal{L}_{\mathrm{top}}$ & Initial-condition and top-boundary losses \\
$L_i^{(0)},\ N_L$ & Warm-up loss component and number of separately weighted loss components \\
$N_d$ & Total number of temperature observations, $N_sN_t$ \\
$N_f,\ N_{\mathrm{top}},\ N_{\mathrm{IC}}$ & Interior, top-boundary, and initial-condition collocation counts \\
$N_h$ & Number of reference-grid points used to evaluate HTC error \\
$N_p$ & Number of boundary control nodes \\
$N_s,\ N_t$ & Number of sensors and measurement times, respectively \\
$\mathcal{N}_h$ & Auxiliary neural network representing the boundary Biot number \\
$\mathbf{p},\ p_j$ & General boundary-parameter vector and component in the adjoint workflow \\
$P$ & PINN early-stopping patience, measured in training iterations \\
$r_J^{(k)},\ r_b^{(k)}$ & Relative objective and boundary-profile changes for adjoint stopping \\
$\mathcal{R}$ & PINN equation, boundary-condition, or initial-condition residual \\
$\mathrm{Re}$ & Jet Reynolds number \\
$\mathrm{RMSE}_h$ & HTC-profile root-mean-square error [$\mathrm{W\,m^{-2}\,K^{-1}}$] \\
$\mathbf{s}_k$ & Boundary quantity monitored by the PINN stopping callback \\
$S_t$ & Target-estimation time saving relative to scratch training [\%] \\
$t,\ t_f$ & Time and final simulation time [$\mathrm{s}$ before nondimensionalization] \\
$t_{\mathrm{scratch}},\ t_{\mathrm{TL}}$ & Production times for scratch-trained and transferred targets [$\mathrm{s}$] \\
$T,\ T_0,\ T_\infty$ & Substrate, initial, and ambient temperatures [$\mathrm{K}$ before nondimensionalization] \\
$T_{\btheta},\mathbf{T}^{n}$ & PINN temperature approximation and discrete state at time step $n$ \\
$\bar{T}_i$ & Mean dimensionless sensor temperature at snapshot $i$ \\
$v$ & Finite-element test function \\
$w_i,\ w_i^{\mathrm{base}}$ & PINN loss weight and its warm-up initialization \\
$w$ & Adjoint stopping-window length in accepted iterations \\
$x,\ y$ & Spatial coordinates [$\mathrm{m}$ before nondimensionalization] \\
$\bar{x}_i,\bar{y}_i$ & Coordinates of sensor $i$ \\
\end{longtable}

\addtocounter{table}{-1}
\begin{longtable}{@{}>{\raggedright\arraybackslash}p{0.23\linewidth}>{\raggedright\arraybackslash}p{0.73\linewidth}@{}}
\textbf{Greek symbols} & \textbf{Definition} \\
\endfirsthead
\textbf{Greek symbols (continued)} & \textbf{Definition} \\
\endhead
$\alpha$ & Thermal diffusivity, $k/(\rho c_p)$ [$\mathrm{m^2\,s^{-1}}$] \\
$\Gamma_c,\Gamma_r$ & Top and right convective boundaries \\
$\gamma_i$ & Sampled multiplier used to refine loss weight $i$ \\
$\delta_{ij}$ & Kronecker delta \\
$\Delta t,\Delta t_{\mathrm{obs}}$ & Forward time step and sampling interval (dimensional intervals in seconds) \\
$\Delta k$ & Number of training iterations between PINN stopping checks \\
$\epsilon$ & Positive denominator safeguard in the adjoint stopping test \\
$\varepsilon_{ik}$ & Measurement perturbation in ~\Cref{eq:measurements} \\
$\epsilon_{ij}$ & Independent standard-normal draw in ~\Cref{eq:noise_model} \\
$\varepsilon_h$ & Relative spatial $L_2$ error of the HTC profile [\%] \\
$\eta$ & Prescribed dimensionless noise fraction \\
$\btheta$ & Trainable temperature-network weights and biases \\
$\blambda,\ \Lambda_j$ & Boundary-parameter vector and its component in the inverse and PINN formulations \\
$\lambda^n$ & Discrete adjoint field associated with the forward step from $n$ to $n+1$ \\
$\mu_{\mathrm{reg}}$ & Adjoint boundary-profile regularization coefficient \\
$\rho$ & Density of the substrate [$\mathrm{kg\,m^{-3}}$] \\
$\sigma_i$ & Spatial standard deviation of dimensionless sensor temperatures at snapshot $i$ \\
$\tau_f,\tau_g$ & L-BFGS-B objective-reduction and projected-gradient tolerances \\
$\tau_{\mathrm{abs}},\tau_{\mathrm{rel}}$ & Absolute and relative PINN boundary-stability tolerances \\
$\phi_j,\psi_j$ & Finite-element boundary basis and PWL hat functions \\
$\omega_i$ & Composite-trapezoidal weight in the boundary-profile norm \\
$\Omega$ & Two-dimensional substrate domain \\
\end{longtable}

\addtocounter{table}{-1}
\begin{longtable}{@{}>{\raggedright\arraybackslash}p{0.23\linewidth}>{\raggedright\arraybackslash}p{0.73\linewidth}@{}}
\textbf{List of Subscripts} & \textbf{Meaning} \\
\endfirsthead
\textbf{List of Subscripts (continued)} & \textbf{Meaning} \\
\endhead
$\mathrm{BC},\mathrm{IC},\mathrm{PDE}$ & Boundary-condition, initial-condition, and governing-equation terms \\
$\mathrm{data},\mathrm{reg}$ & Data mismatch and regularization \\
$f$ & Final time in $t_f$; interior collocation points in $N_f$ \\
$fc,\ fr$ & Forced convection at the top and free convection at the right boundary \\
$\mathrm{obs},\mathrm{pred}$ & Observed and predicted temperatures \\
\end{longtable}

\addtocounter{table}{-1}
\begin{longtable}{@{}>{\raggedright\arraybackslash}p{0.23\linewidth}>{\raggedright\arraybackslash}p{0.73\linewidth}@{}}
\textbf{Superscripts and accents} & \textbf{Meaning} \\
\endfirsthead
\textbf{Superscripts and accents (continued)} & \textbf{Meaning} \\
\endhead
$\mathrm{est},\mathrm{ref}$ & Estimated and reference HTC profiles \\
$\mathrm{meas},\mathrm{obs}$ & Measured or observed temperatures \\
$\mathrm{poly},\mathrm{PWL},\mathrm{NN}$ & Polynomial, piecewise-linear, and neural-network boundary representations \\
\end{longtable}

\addtocounter{table}{-1}
\begin{longtable}{@{}>{\raggedright\arraybackslash}p{0.23\linewidth}>{\raggedright\arraybackslash}p{0.73\linewidth}@{}}
\textbf{Acronyms} & \textbf{Definition} \\
\endfirsthead
\textbf{Acronyms (continued)} & \textbf{Definition} \\
\endhead
BC & Boundary condition \\
CFD & Computational fluid dynamics \\
CHT & Conjugate heat transfer \\
FEM & Finite element method \\
HTC & Heat transfer coefficient \\
IC & Initial condition \\
L-BFGS-B & Limited-memory Broyden--Fletcher--Goldfarb--Shanno algorithm with bound constraints \\
NN & Neural network \\
PDE & Partial differential equation \\
PINN & Physics-informed neural network \\
PWL & Piecewise linear \\
RMSE & Root-mean-square error \\
TL & Transfer learning \\
TPE & Tree-structured Parzen estimator \\
\end{longtable}
\endgroup


\section{Introduction}
\label{sec:introduction}
Efficient thermal management is crucial for numerous engineering applications, including gas turbine blade cooling \citep{han_recent_2004}, aircraft protection~\citep{zhao_inverse_2025}, high-power electronics \citep{ventola_rough_2014}, metal processing \citep{sengupta_use_2005}, additive manufacturing \citep{hackenhaar_experimentalnumerical_2020}, and energy systems \citep{zhao_review_2021}.
Among the different cooling approaches, jet impingement cooling is widely employed in thermal management because it
maintains high local heat-transfer rates over small, confined target areas
\citep{martin_heat_1977,zuckerman_jet_2006}. The thermal performance of such
systems is governed primarily by the spatial distribution of the convective
heat transfer coefficient (HTC). Accurate knowledge of this distribution is
therefore essential for thermal design, cooling-system assessment, model
calibration, and digital twins for monitoring and control~\citep{xue_realtime_2026}.
However, direct measurement of the local HTC is generally impractical,
motivating inverse techniques that estimate the boundary coefficient from
sparse and noisy temperature measurements obtained at interior locations within the substrate.

Inverse heat transfer problems have traditionally been formulated as PDE-constrained optimization problems~\citep{hinze_optimization_2009} in which the unknown boundary conditions or material properties are estimated by minimizing the data-misfit between measured and predicted temperatures. Due to the ill-posed nature of these problems, stable estimation requires appropriate regularization together with efficient optimization algorithms~\citep{alifanov_inverse_1994,ozisik_inverse_2021}. In recent decades, numerous deterministic methods--such as gradient-based optimization~\citep{huang_threedimensional_1999}, sequential estimation methods~\citep{beck_nonlinear_1970}, regularization techniques ~\citep{tikhonov_solutions_1977,engl_regularization_1996}, and discrete adjoint finite element methods~\citep{jarny_general_1991,giles_introduction_2000}--have been effectively used for boundary parameter estimation. Despite their notable performance, performing repeated inverse analyses across varying operating conditions is still computationally intensive, especially for design optimization, repeated parametric investigations, and digital twin applications.

Physics-informed neural networks (PINNs) have recently emerged as a promising alternative for  PDE-constrained inverse problems. By embedding the governing equations into the loss function via automatic differentiation, PINNs enforce physical consistency while assimilating sparse measurement data~\citep{raissi_physicsinformed_2019,cuomo_scientific_2022}. As a result, PINNs have been used successfully in a range of inverse heat transfer applications, such as identifying thermal conductivity, heat sources, and unknown boundary conditions~\citep{cai_physicsinformed_2021,he_physicsinformed_2021}. However, most current PINN-based inverse approaches still require training a new neural network from scratch for each new measurement dataset under different operating conditions, making iterative design workflows that demand repeated analyses computationally costly.

Previously, we developed a physics-informed neural network (PINN)-based framework for inverse  estimation of the boundary convective heat transfer coefficient in jet impingement cooling using sparse transient temperature measurements. The proposed framework was validated against three-dimensional computational fluid dynamics(CFD)-conjugate heat transfer (CHT) simulations and demonstrated robustness under significant multiplicative measurement noise. To reduce the dimensionality of the inverse problem, the unknown boundary coefficient was represented using a low-order polynomial parameterization. 
This representation provided a sufficiently compact inverse formulation as the assumed functional family was consistent with the underlying boundary heat transfer coefficient~\citep{hazra_pinnbased_2026a}. 
it implicitly prescribed a global shape for the boundary profile a priori, which may limit the recovery of more complex or previously unseen spatial variations. To broaden the inverse framework to  more general boundary heat-transfer profiles, we require parameterization approaches that provide greater representational flexibility while maintaining stability with sparse temperature measurements.

To overcome these limitations, we consider two more flexible alternatives. The first uses a piecewise-linear function with trainable nodal values, while the second represents the boundary coefficient with an auxiliary neural network. Both allow the HTC profile to vary without prescribing a quadratic form, although the increased number of parameters also makes the inverse problem more expensive to solve. This additional cost matters when inverse estimation has to be performed repeatedly. Such repeated analyses arise, for example, when operating conditions or measurement data change during design studies, uncertainty quantification, or system monitoring. Training a new PINN from scratch for every case can then become expensive. This raises a more practical question: can information learned from one inverse problem be reused when solving another related problem?

Transfer learning provides a natural mechanism for reusing information learned from prior, related inverse problems. Earlier work has presented transfer-learning approaches for PINN-based inverse problems when loading conditions or unknown coefficients vary, reporting lower training costs or better convergence~\citep{xu_transfer_2023,lin_gradientenhanced_2024}. In addition, transfer learning has been investigated for heat-transfer problems, such as effusion cooling~\citep{yu_transfer_2023} and phase-change applications~\citep{jalili_transfer_2024}. However, its use for repeated inverse estimation of a spatially distributed thermal boundary coefficient under changes in measurement noise, Reynolds number, and boundary-profile characteristics remains less explored.
In this work, transfer learning is investigated across two source--target configurations. The first considers transfer across different noise levels while keeping the Reynolds number and boundary-profile family fixed. The second examines transfer from a controlled Gaussian-profile inverse problem to CHT-based inverse problems at varying Reynolds numbers. These two setups allow us to examine how well transfer learning reduces computational cost while preserving reconstruction accuracy across different source--target relationships.

Although PINNs provide a flexible framework for inverse heat-transfer problems, discrete adjoint finite element methods~\citep{giles_introduction_2000,jarny_general_1991} are mature, widely used techniques for PDE-constrained parameter estimation. Accordingly, the proposed PINN and Adjoint-FEM formulations are compared in a systematic manner, focusing on reconstruction accuracy and wall-clock time under the same inverse setup and measurement configuration. Transfer learning is assessed separately with respect to how effectively it can lower the computational cost of repeated PINN-based inverse estimation. Instead of aiming to identify a single best method overall, the goal is to describe and contrast how the two approaches behave for the inverse heat-transfer cases considered.

The proposed framework is assessed using controlled benchmark problems with prescribed boundary profiles and high-fidelity CFD-CHT-generated cases for jet impingement cooling~\citep{hazra_pinnbased_2026a}. The controlled benchmarks allow a systematic evaluation of the boundary parameterization and transfer-learning strategies, whereas the CFD-CHT-based cases supply independently produced reference solutions from a standard high-fidelity computational workflow and illustrate the applicability of the proposed inverse framework to the intended engineering problem. In addition, a discrete-adjoint finite element method is used as an independent inverse approach to evaluate reconstruction accuracy and computational efficiency.

The main contributions of this study are:
\begin{itemize}
	\item Design of two general boundary-parameterization approaches--a piecewise 
linear model and a neural-network-based predictor--for inverse 
reconstruction of spatially varying convective heat transfer coefficients.
	\item Formulation and systematic assessment of transfer-learning methods to speed 
up repeated PINN-based inverse analyses across different 
boundary-profile classes and measurement-noise conditions.
	\item Benchmarking PINN-based inverse inference against a separate discrete Adjoint-FEM framework with respect to HTC reconstruction accuracy and overall computational expense using the same observation targets.
	\item The proposed inverse formulations are evaluated against high-fidelity three-dimensional CFD--CHT-derived HTC profiles across multiple Reynolds numbers.
\end{itemize}

The remainder of the paper is organized as follows. \Cref{sec:problem} presents the physical problem and the boundary parameterizations used for inverse estimation. \Cref{sec:methodology} describes the PINN formulation, transfer-learning strategy, and discrete Adjoint-FEM approach. \Cref{sec:setup} details the benchmark datasets, sensor configuration,
noise model, numerical implementation, and performance metrics. \Cref{sec:results} compares the boundary representations and inverse methods, evaluates transfer learning across measurement-noise levels and boundary-profile families. Section~\Cref{sec:conclusions} summarizes the main findings. The appendices contain additional information on adjoint regularization, L-curve diagnostics, production stopping criteria, and convergence histories.


\section{Problem Formulation}
\label{sec:problem}

\subsection{Physical problem}
\label{subsec:model}
\begin{figure}[!ht]
  \centering
  \resizebox{\linewidth}{!}{%
  \includegraphics{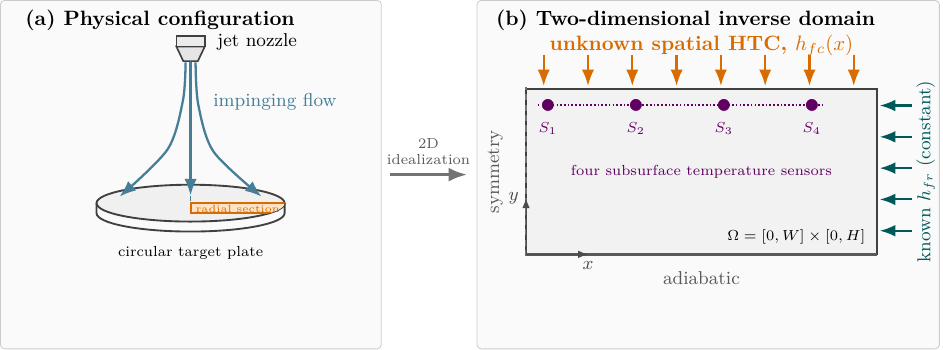}
  }
  \caption{Physical setting and reduced inverse-analysis domain.
  \textbf{(a)} An axisymmetric impinging jet cools a circular plate; the
  highlighted radial section motivates the two-dimensional model.
  \textbf{(b)} The reduced domain contains an unknown spatial HTC
  $h_{fc}(x)$, a known constant lateral coefficient $h_{fr}$, symmetry
  and adiabatic boundaries, and the four subsurface temperature sensors
  shared by all inverse reconstructions. The geometric reduction follows
  the physical idealization in our previous study
  \citep{hazra_pinnbased_2026a}, while the schematic is redrawn for the
  present generalized representations and measurement configuration.}
  \label{fig:problem-domain}
\end{figure}

The present work considers the inverse estimation of a spatially varying boundary heat transfer coefficient (HTC) from transient temperature measurements acquired within a cooled substrate. The physical system consists of an axisymmetric jet impinging normally on a circular plate and is inherently governed by three-dimensional conjugate heat transfer. Exploiting the axisymmetry of the normally impinging jet and the circular substrate, the inverse analysis is performed using a reduced two-dimensional transient heat conduction model representing the radial cross-section of the substrate, as illustrated in \Cref{fig:problem-domain}. The left panel of \Cref{fig:problem-domain} presents a schematic of the three-dimensional jet impingement cooling configuration, whereas the right panel shows the corresponding two-dimensional computational domain employed as the forward model for inverse parameter estimation. The accuracy of the reduced inverse model for boundary parameter estimation was established in our previous work through detailed comparison with full three-dimensional computational fluid dynamics (CFD)-conjugate heat transfer (CHT) simulations~\cite{hazra_pinnbased_2026a}.

The present study uses HTC distributions and transient temperature fields
generated from the three-dimensional CFD--CHT simulations as high-fidelity
reference data. Synthetic temperature measurements are extracted from the
CFD--CHT temperature fields at prescribed sensor locations and observation
times and supplied to the inverse methods, whereas the corresponding
CFD--CHT HTC distributions are used to assess the reconstructed boundary
profiles. Both the proposed PINN formulations and the discrete Adjoint-FEM
method solve the same reduced two-dimensional inverse problem. The detailed
three-dimensional CFD--CHT model is therefore not repeated here; instead,
the following formulation defines the reduced forward model used throughout
the inverse analyses.

 Within the reduced formulation, the thermal influence of the impinging jet is incorporated through a spatially varying convective boundary condition prescribed on the cooled surface, while heat conduction within the substrate is governed by the transient heat conduction equation:
\begin{align}
\rho c_p\,\pd{T}{t} = k\left(\pdd{T}{x}+\pdd{T}{y}\right),
\quad (x,y)\in\Omega,\; t\in(0,t_f],
\label{eq:governing}
\end{align}
\noindent where $T(x,y,t)$ denotes the temperature field within the substrate; $\rho$, $c_p$, and $k$ are the density, specific heat capacity, and thermal conductivity,
respectively, assumed known and constant.

The governing equation is supplemented by the following boundary conditions:
\begin{align}
-k\pd{T}{y}\bigg|_{y=H} &= h_{fc}(x)\left[T(x,H,t)-T_{\infty}\right],
&&\text{(top: jet-impingement boundary)}, \label{eq:bc_top}\\[6pt]
-k\pd{T}{x}\bigg|_{x=W} &= h_{fr}\left[T(W,y,t)-T_{\infty}\right],
&&\text{(right: lateral cooling boundary)}, \label{eq:bc_right}\\[6pt]
\pd{T}{x}\bigg|_{x=0} &= 0,
&&\text{(left: symmetry)}, \label{eq:bc_left}\\[6pt]
\pd{T}{y}\bigg|_{y=0} &= 0,
&&\text{(bottom: adiabatic)}. \label{eq:bc_bottom}
\end{align}

Here, $h_{fc}(x)$ is the spatially varying HTC at the jet-facing boundary,
$h_{fr}$ is the fixed uniform HTC at the right boundary, and $T_{\infty}$ is
the ambient temperature. 

The initial temperature field is prescribed as

\begin{align}
T(x,y,0)=T_0, \quad (x,y)\in\overline{\Omega},
\label{eq:ic}
\end{align}
\noindent where $T_0$ is known.

We introduce the following dimensionless variables:
\begin{equation}
	\tilde{x}=\frac{x}{W},\quad
	\tilde{y}=\frac{y}{W},\quad
	\tilde{t}=\frac{\alpha t}{W^2},\quad
	\tilde{T}=\frac{T-T_{\infty}}{T_0-T_{\infty}},
	\label{eq:nondim_vars}
\end{equation}

\noindent where $\alpha=k/(\rho c_p)$ is the thermal diffusivity. 
The governing equations can then be written in the following nondimensional form; for simplicity, we drop the tildes and subsequently use the same symbols to represent the dimensionless variables:

\begin{equation}
	\pd{T}{t}=\pdd{T}{x}+\pdd{T}{y},
	\quad (x,y)\in[0,1]\times[0,L_r],\; t\in[0,t_f],
	\label{eq:governing_nondim}
\end{equation}
with initial and boundary conditions
\begin{align}
	T(x,y,0)&=1, \notag\\
	\pd{T}{y}(x,L_r,t)+\Bi_{fc}(x)\,T&=0,
	\quad \pd{T}{y}(x,0,t)=0, \notag\\
	\pd{T}{x}(1,y,t)+\Bi_{fr}\,T&=0,
	\quad \pd{T}{x}(0,y,t)=0,
	\label{eqn:heat_non_dim_aux_con}
\end{align}
where
\begin{align} 
	\Bi_{fc}(x)=\frac{h_{fc}(x)W}{k},\quad
	\Bi_{fr}=\frac{h_{fr}W}{k},\quad
	L_r=\frac{H}{W}.
	\label{eq:nondim_groups}
\end{align}

Here, \(\Bi_{fc}(x)\) and \(\Bi_{fr}\) are the Biot numbers associated with convection at the top and right boundaries, respectively. 
The quantity of engineering interest is the spatially varying convective heat transfer coefficient $h(x)$. The Biot number is introduced solely as a nondimensional variable for mathematical formulation of the inverse problem; because the characteristic length and thermal conductivity are known, reconstruction of the Biot number is mathematically equivalent to reconstruction of the corresponding heat transfer coefficient. Accordingly, the inverse formulation is expressed in terms of the Biot number, whereas all results are presented as dimensional heat transfer coefficients to facilitate comparison with CFD-CHT data.

\subsection{Inverse problem statement and representation strategy}
\label{subsec:inverse_formulation}

The objective of the current inverse problem is to reconstruct the unknown spatially varying Biot number $Bi_{fc}(x)$ using transient temperature measurements acquired at a set of discrete points inside the computational domain. These measurements are linked to the forward-model solution by
\begin{equation}
T^{\mathrm{meas}}_{ik}=T(\bar{x}_i,\bar{y}_i,t_k)+\varepsilon_{ik},
\label{eq:measurements}
\end{equation}
where $N_s$ is the number of interior sensors located at $\{(\bar{x}_i,\bar{y}_i)\}_{i=1}^{N_s}\subset\Omega$, each recording temperatures at the non-dimensional time instants $\{t_k\}_{k=1}^{N_t}\subset(0,t_f]$ and $\varepsilon_{ik}$ denoting the measurement noise. 

The boundary parameter is inferred by minimizing the discrepancy between
the measured and model-predicted temperatures. For a given boundary
parameterization $\blambda$, the data-misfit functional is
defined as
\begin{equation}
\mathcal{J}_{\mathrm{data}}(\blambda)
=
\frac{1}{N_sN_t}
\sum_{i=1}^{N_s}
\sum_{k=1}^{N_t}
\left[
T(\bar{x}_i,\bar{y}_i,t_k;\blambda)
-
T^{\mathrm{meas}}_{ik}
\right]^2,
\label{eq:data_misfit}
\end{equation}
where $\blambda$ denotes the parameters defining the unknown
boundary Biot number. The inverse problem is therefore to determine
$\blambda$ such that the model predictions are consistent with
the available temperature measurements.

Because $\Bi_{fc}(x)$ is a spatially distributed unknown, it is represented
using a finite-dimensional parameterization for numerical estimation. Three representations are considered: (i) A low-dimensional quadratic parameterization adopted from our previous work~\citep{hazra_pinnbased_2026a}, (ii) A piecewise-linear (PWL) parameterization, and (iii) A neural-network (NN) parameterization. The quadratic and PWL representations are used in both the PINN and Adjoint-FEM formulations, whereas the NN representation is considered only within the PINN framework. The corresponding optimization procedures are described in \Cref{sec:methodology}.

As the low-dimensional representation, the unknown boundary parameter is represented using a quadratic polynomial. Following our previous PINN formulation,
\begin{align}
\Bi_{fc}^{\mathrm{poly}}(x;\mathbf{c})
&=
\sum_{r=0}^{2} c_r x^r,
\qquad
\mathbf{c}=
[c_0,c_1,c_2]^{\mathsf T}.
\label{eq:polynomial_bi}
\end{align}
This parameterization reduces the distributed inverse problem to a small set of global coefficients and is computationally efficient when the underlying boundary profile is well described by the assumed polynomial family. Its limitation is structural, since spatial features that cannot be represented by the prescribed polynomial form cannot be recovered.

To relax this prescribed functional form, the boundary field is
parameterized through nodal values using piecewise-linear interpolation. Let $N_p$ control nodes be distributed uniformly over the
nondimensional boundary interval,
\begin{equation}
x_j=\frac{j}{N_p-1},
\qquad
j=0,\ldots,N_p-1.
\label{eq:pwl_nodes}
\end{equation}
The unknown profile is then expressed as
\begin{align}
\Bi_{fc}^{\mathrm{PWL}}(x;\mathbf{b})=
\sum_{j=0}^{N_p-1} b_j\psi_j(x),
\qquad
\mathbf{b}=
[b_0,\ldots,b_{N_p-1}]^{\mathsf T},
\label{eq:pwl_bi}
\end{align}
where $\psi_j(x)$ are the piecewise-linear hat functions. Since
$\psi_j(x_i)=\delta_{ij}$, $b_j$ represents the Biot number at the
corresponding control node. The compact support of the basis functions
provides local control of the profile, while avoiding a prescribed global
analytical shape.

A neural-network (NN) parameterization is also considered, in which the
boundary coordinate is mapped directly to the Biot number,
\begin{equation}
\Bi_{fc}^{\mathrm{NN}}(x)
=
\mathcal{N}_{h}(x),
\label{eq:nn_bi}
\end{equation}
where $\mathcal{N}_{h}$ denotes a neural-network function that parameterizes
the spatially varying boundary Biot number. This representation does not
prescribe a specific analytical form or polynomial order for the boundary
profile.

%

\section{Inverse Estimation Methodology}
\label{sec:methodology}

The inverse problem formulated in \Cref{subsec:inverse_formulation} is solved using two complementary approaches: a physics-informed neural network (PINN) and a discrete Adjoint-FEM method. The PINN formulation and its transfer-learning extension are described first, followed by the Adjoint-FEM formulation.

\subsection{PINN-based Inverse Estimation}
\label{subsec:pinn}
\subsubsection{PINN Formulation}
\label{sec:pinn_formulation}

In the PINN formulation, the dimensionless transient temperature field is
approximated by a fully connected neural network
$T_{\btheta}(x,y,t)$, where $\btheta$ denotes the trainable network
parameters. The parameters defining the unknown boundary representation are
denoted collectively by $\blambda$, as introduced in
\Cref{subsec:inverse_formulation}. Thus, the temperature-network parameters
and the boundary parameters are optimized simultaneously.

The governing heat-conduction equation is enforced at $N_f$ interior
collocation points through the residual
\begin{align}
	\mathcal{R}_{\mathrm{PDE}}
	&=
	\frac{\partial T_{\btheta}}{\partial t}
	-
	\frac{\partial^2 T_{\btheta}}{\partial x^2}
	-
	\frac{\partial^2 T_{\btheta}}{\partial y^2}.
	\label{eq:pde_residual}
\end{align}
Here, $N_f$ denotes the number of interior collocation points. The
corresponding PDE loss is defined as
\begin{align}
	\mathcal{L}_{\mathrm{PDE}}
	&=
	\frac{1}{N_f}
	\sum_{q=1}^{N_f}
	\left(
	\mathcal{R}_{\mathrm{PDE},q}
	\right)^2.
	\label{eq:pde_loss}
\end{align}

The unknown convective boundary condition at $y=L_r$ is imposed through
the residual
\begin{align}
	\mathcal{R}_{\mathrm{top}}
	&=
	\frac{\partial T_{\btheta}}{\partial y}
	+
	\Bi_{fc}(x;\blambda)T_{\btheta},
	\qquad y=L_r.
	\label{eq:top_bc_residual}
\end{align}
For $N_{\mathrm{top}}$ collocation points sampled on the top boundary, the
corresponding loss is
\begin{align}
	\mathcal{L}_{\mathrm{top}}
	&=
	\frac{1}{N_{\mathrm{top}}}
	\sum_{q=1}^{N_{\mathrm{top}}}
	\left(
	\mathcal{R}_{\mathrm{top},q}
	\right)^2.
	\label{eq:top_loss}
\end{align}

The residuals associated with the remaining boundary conditions are
\begin{align}
	\mathcal{R}_{\mathrm{left}}
	&=
	\frac{\partial T_{\btheta}}{\partial x},
	\qquad x=0,
	\\
	\mathcal{R}_{\mathrm{right}}
	&=
	\frac{\partial T_{\btheta}}{\partial x}
	+
	\Bi_{fr}T_{\btheta},
	\qquad x=1,
	\\
	\mathcal{R}_{\mathrm{bottom}}
	&=
	\frac{\partial T_{\btheta}}{\partial y},
	\qquad y=0.
	\label{eq:known_bc_residuals}
\end{align}
The corresponding mean-squared losses are
\begin{align}
	\mathcal{L}_{\mathrm{left}}=
	\frac{1}{N_{\mathrm{left}}}
	\sum_{q=1}^{N_{\mathrm{left}}}
	\left(
	\mathcal{R}_{\mathrm{left},q}
	\right)^2,
	\quad
	\mathcal{L}_{\mathrm{right}}=
	\frac{1}{N_{\mathrm{right}}}
	\sum_{q=1}^{N_{\mathrm{right}}}
	\left(
	\mathcal{R}_{\mathrm{right},q}
	\right)^2,
	\quad
	\mathcal{L}_{\mathrm{bottom}}=
	\frac{1}{N_{\mathrm{bottom}}}
	\sum_{q=1}^{N_{\mathrm{bottom}}}
	\left(
	\mathcal{R}_{\mathrm{bottom},q}
	\right)^2.
	\label{eq:known_bc_losses}
\end{align}

The initial condition is enforced through
\begin{align}
	\mathcal{R}_{\mathrm{IC}}
	&=
	T_{\btheta}(x,y,0)-1,
	\label{eq:ic_residual}
\end{align}
with the corresponding loss evaluated at $N_{\mathrm{IC}}$
initial-condition collocation points:
\begin{align}
	\mathcal{L}_{\mathrm{IC}}=
	\frac{1}{N_{\mathrm{IC}}}
	\sum_{q=1}^{N_{\mathrm{IC}}}
	\left(
	\mathcal{R}_{\mathrm{IC},q}
	\right)^2.
	\label{eq:ic_loss}
\end{align}

The temperature measurements are incorporated through the data-mismatch
loss. For $N_s$ sensor locations and $N_t$ measurement times, the total
number of observations is $N_d=N_sN_t$, and
\begin{align}
	\mathcal{L}_{\mathrm{data}}
	&=
	\frac{1}{N_d}
	\sum_{i=1}^{N_s}
	\sum_{k=1}^{N_t}
	\left[
	T_{\btheta}(\bar{x}_i,\bar{y}_i,t_k)
	-
	T^{\mathrm{meas}}_{ik}
	\right]^2.
	\label{eq:pinn_data_loss}
\end{align}

The individual loss components are combined to form the total PINN
objective,
\begin{align}
	\mathcal{L}
	&=
	w_{\mathrm{PDE}}\mathcal{L}_{\mathrm{PDE}}
	+
	w_{\mathrm{top}}\mathcal{L}_{\mathrm{top}}
	+
	w_{\mathrm{left}}\mathcal{L}_{\mathrm{left}}
	+
	w_{\mathrm{right}}\mathcal{L}_{\mathrm{right}}
	+
	w_{\mathrm{bottom}}\mathcal{L}_{\mathrm{bottom}}
	+
	w_{\mathrm{IC}}\mathcal{L}_{\mathrm{IC}}
	+
	w_{\mathrm{data}}\mathcal{L}_{\mathrm{data}},
	\label{eq:total_pinn_loss}
\end{align}
where the positive coefficients $w_i$ control the relative contribution
of the individual residual components to the optimization objective.

Because the magnitudes of the individual loss components can differ
substantially, the weights are initialized according to the relative
magnitudes of the corresponding loss components ~\citep{wang_understanding_2021}. Following a short
unit-weight warm-up, let $L_i^{(0)}$ denote the value of the $i$th loss
component. Its initial weight is defined relative to the mean loss
magnitude as
\begin{align}
	w_i^{\mathrm{base}}
	&=
	\operatorname{clip}\!\left(
	\frac{L_i^{(0)}}
	{\displaystyle
		\frac{1}{N_L}\sum_{j=1}^{N_L}L_j^{(0)}},
	w_{\min},w_{\max}
	\right),
	\qquad
	N_L=7,
	\label{eq:base_loss_weight}
\end{align}
where $N_L$ is the number of separately weighted loss components and
$w_{\min}$ and $w_{\max}$ denote prescribed lower and upper bounds.
The resulting weights are subsequently refined through offline
hyperparameter selection, as described in
\Cref{subsec:pinn_setup}.

The temperature-network parameters and the boundary parameters are then
optimized jointly according to
\begin{align}
	(\btheta^{*},\blambda^{*})
	&=
	\underset{\btheta,\blambda}{\operatorname{arg\,min}}
	\;
	\mathcal{L}(\btheta,\blambda).
	\label{eq:pinn_optimization}
\end{align}

\subsubsection{Boundary-Parameter Optimization}
\label{sec:boundary_parameter_optimization}

The boundary parameters $\blambda$ are optimized jointly with the temperature-network parameters $\btheta$ through the PINN loss. For the top-boundary residual, the direct contribution of the boundary loss to the gradient with respect to a boundary parameter $\Lambda_j$ is
\begin{align}
\frac{\partial\mathcal{L}_{\mathrm{top}}}
     {\partial\Lambda_j}
&=
\frac{2}{N_{\mathrm{top}}}
\sum_{q=1}^{N_{\mathrm{top}}}
\mathcal{R}_{\mathrm{top},q}\,
T_{\btheta,q}\,
\frac{\partial\Bi_{fc}(x_q;\blambda)}
     {\partial\Lambda_j},
\label{eq:pinn_boundary_gradient}
\end{align}
up to the corresponding loss weight. Thus, the boundary parameterization determines the sensitivity of the boundary residual to each trainable parameter.

For the PWL representation,
\begin{align}
\frac{\partial\Bi_{fc}^{\mathrm{PWL}}}
     {\partial b_j}
&=
\psi_j(x),
\label{eq:pwl_parameter_gradient}
\end{align}
so the direct sensitivity of a boundary collocation point to $b_j$ is nonzero only where the corresponding basis function has support. In contrast, for the auxiliary boundary network, the chain rule propagates the boundary-residual gradient through the network parameters, resulting in parameter updates that are generally coupled across the spatial domain.

The boundary parameters are also affected indirectly by the data, PDE, initial-condition, and known-boundary losses through their dependence on the temperature network. Consequently, the final boundary estimate results from the joint optimization of the temperature field and the boundary representation.
\subsubsection{Transfer learning}
\label{subsubsec:transfer_learning}
Transfer learning exploits information learned from a source inverse problem
to facilitate the solution of a related target inverse problem. In the present
framework, a trained source PINN provides an informed initialization of the
temperature-field representation for the target problem. The target boundary
parameters remain adaptable, allowing the transferred model to accommodate
changes in the observations and operating conditions. By reusing the transferable components of the source model,
transfer learning has the potential to reduce the computational effort
required for target inversion compared with training a PINN from scratch.

The benefit of transfer learning depends on the similarity between the
source and target problems. A source model trained on a closely related
problem is expected to provide a more useful initialization than one trained
on a substantially different problem. In the present study, transfer is performed between models using the same boundary representation. Specifically, PWL source models are transferred to
target PWL models, while NN-estimator source models are transferred to
target NN-estimator models. For NN-estimator-to-NN-estimator transfer, the
source and target auxiliary networks used for boundary-parameter estimation
have the same architecture, allowing the corresponding pretrained network
weights to be transferred to the target model. The target boundary
parameters remain trainable during subsequent optimization. The source and
target inverse problems may otherwise differ in their observations,
measurement noise, Reynolds number of the impinging jet, and underlying boundary profile.

This study examines two transfer settings: (i) transfer within the same physical problem, but using observation datasets corrupted by varying noise levels, and (ii) transfer across boundary profiles and jets at different Reynolds numbers. The first setting changes only the measurement noise, whereas the second differs in boundary profile, Reynolds number, and data-generation setting.

\begin{figure}[htbp]
\centering
\resizebox{1.12\textwidth}{!}{%
\includegraphics{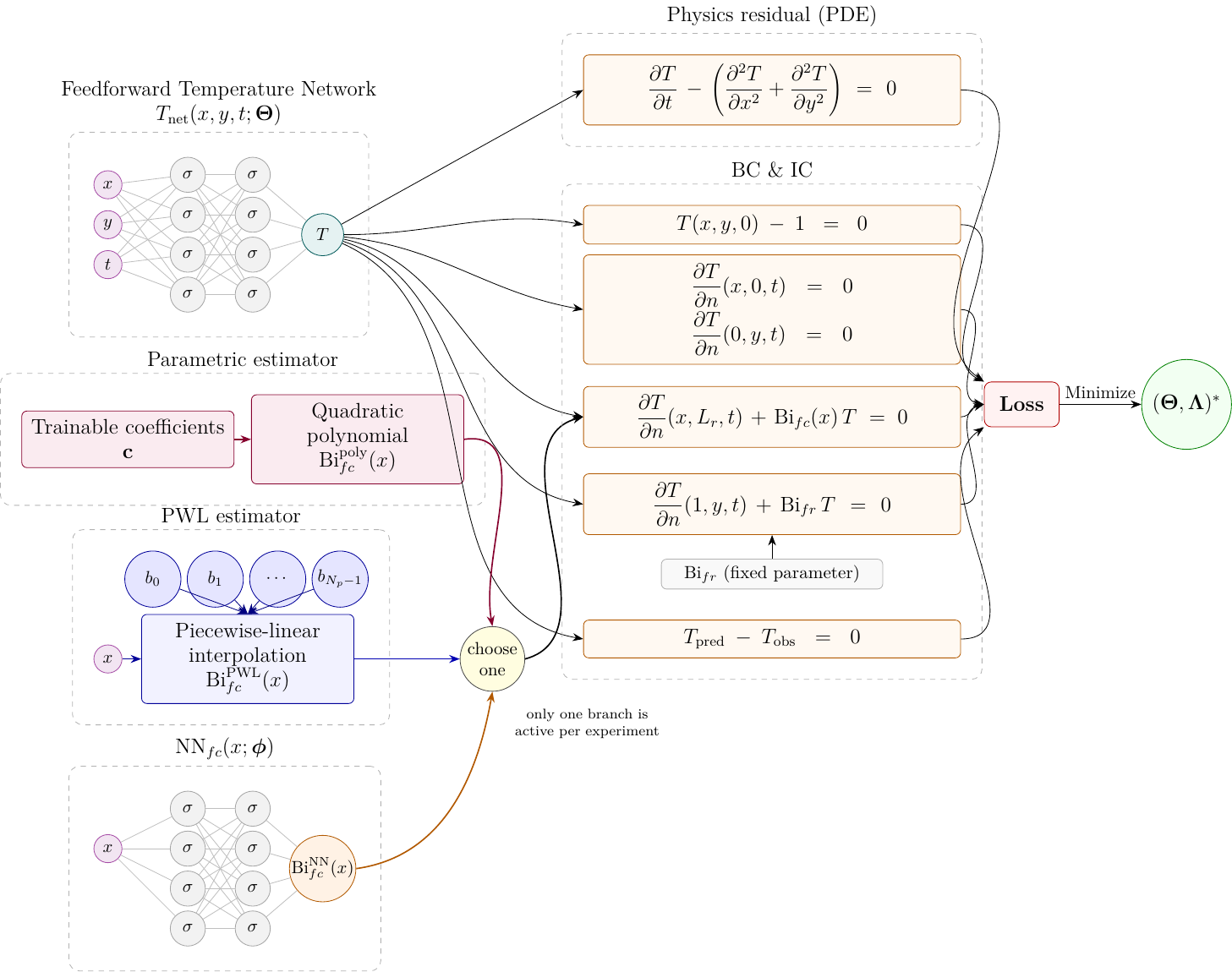}%
}
\caption{Architecture of the PINN inverse framework. The primary network
$T_{\mathrm{net}}(x,y,t;\btheta)$ predicts the temperature field $T$, which feeds the
PDE residual and each boundary/initial condition (BC/IC) loss term. The
spatially varying top-boundary Biot number $\Bi_{fc}(x)$ is supplied by one
of three alternative estimator branches: a low-dimensional quadratic
parametric profile, a piecewise-linear (PWL) control-node
parameterization, or an auxiliary
neural network $\mathrm{NN}_{fc}(x;\boldsymbol{\phi})$. Each branch enters only
through the top Robin boundary condition, while the right boundary uses a fixed
known Biot number $\Bi_{fr}$. All loss terms are summed into a single total
objective and minimized jointly over the temperature-network parameters
$\btheta$ and collective boundary-estimator parameters $\blambda$. Depending
on the active branch, $\blambda$ denotes the parametric coefficients
$\mathbf{c}$, PWL nodal values $\mathbf{b}$, or NN weights and biases
$\boldsymbol{\phi}$.}
\label{fig:pinn_architecture}
\end{figure}

\subsection{Adjoint-FEM formulation}
\label{subsec:adjoint}

The same inverse problem is solved independently using a discrete adjoint
finite element method (Adjoint-FEM). In contrast to the PINN formulation,
the temperature field is obtained from a discretized forward finite element
solve, while the discrete adjoint equations provide the gradient of the
inverse objective with respect to the boundary parameters. The Adjoint-FEM
therefore provides an independent gradient-based methodology for comparison
with the proposed PINN framework.

Let $V_h\subset H^1(\Omega)$ be a conforming finite-element space
\citep{brenner_mathematical_2008,ern_theory_2004}. Using a backward Euler
discretization in time, the weak form of the forward problem is
\begin{align}
  \int_\Omega
  \frac{T^{n+1}-T^n}{\Delta t}v\,\dO
  +\int_\Omega\nabla T^{n+1}\cdot\nabla v\,\dO
  +\int_{\Gc}\Bi_{fc}T^{n+1}v\,\dG
  +\int_{\Gamma_r}\Bi_{fr}T^{n+1}v\,\dG=0,
  \label{eq:weak_form}
\end{align}
for all $v\in V_h$, where $\Gc$ and $\Gamma_r$ denote the top and
right convective boundaries, respectively.
After finite-element assembly,
the forward problem at time level $n+1$ can be written as
\begin{align}
  \mathbf{A}\mathbf{T}^{n+1}
  =\Delta t^{-1}\mathbf{M}\mathbf{T}^{n},
  \qquad
  \mathbf{A}
  =\Delta t^{-1}\mathbf{M}
  +\mathbf{K}
  +\mathbf{B}_{fc}
  +\mathbf{B}_{fr}.
  \label{eq:forward_matrix}
\end{align}
where $\mathbf{M}$ and $\mathbf{K}$ are the mass and stiffness matrices,
respectively, while $\mathbf{B}_{fc}$ and $\mathbf{B}_{fr}$ represent the
Robin contributions at the top and right boundaries.

The discrete inverse objective is defined as

\begin{align}
\mathcal{J}(\Bi_{fc})
=
\frac{1}{2}
\sum_{k\in\mathcal{K}_{\mathrm{obs}}}
\left\|
\mathbf{H}\mathbf{T}^{k}
-
\mathbf{T}^{\mathrm{obs}}_k
\right\|^2
+
\mathcal{J}_{\mathrm{reg}}(\Bi_{fc}),
\label{eq:cost_functional}
\end{align}

where $\mathbf{H}$ maps the finite-element solution to the observation
locations and $\mathcal{K}_{\mathrm{obs}}$ denotes the set of observation
times. 

For the generalized boundary reconstruction, the unknown Biot number is
represented using a piecewise-linear parameterization,
\begin{align}
  \Bi_{fc}(x;\mathbf{b})
  =
  \sum_{j\in\mathcal{D}_{\Gc}}b_j\phi_j(x),
  \qquad x\in\Gc.
  \label{eq:nodal_bi_fem}
\end{align}
where $b_j$ are the boundary control parameters and $\phi_j(x)$ are the
corresponding basis functions on $\Gamma_c$. The boundary values
$\mathbf{b}$ therefore constitute the control variables of the inverse
problem.

For the generalized PWL reconstruction, a first-order smoothness
regularizer is employed to mitigate unsupported spatial oscillations through the $H^1$ seminorm~\citep{jin_error_2021},
\begin{align}
	\mathcal{J}_{\mathrm{reg}}
	&=
	\frac{\mu_{\mathrm{reg}}}{2}
	\int_{\Gamma_c}
	\left|
	\frac{\partial\Bi_{fc}}{\partial x}
	\right|^2
	\,\dG.
	\label{eq:adjoint_regularization}
\end{align}
Here, $\mu_{\mathrm{reg}}$ is the regularization coefficient. The
regularization penalizes spatial gradients of the boundary profile and
does not include a zeroth-order $L^2$ penalty or a reference-profile
prior. The treatment of the PWL representation, coefficient bounds,
regularization selection, and optimization stopping criteria is given
in \Cref{app:adjoint_fem}.

The dependence of the forward solution on the boundary control is
propagated to the objective through a discrete adjoint solve. The
forward recursion in \Cref{eq:forward_matrix} is enforced through the
discrete Lagrangian
\begin{align}
\mathcal{L}
=
\mathcal{J}
-
\sum_{n=0}^{N-1}
\left(\boldsymbol{\lambda}^{n}\right)^{\top}
\left(
\mathbf{A}\mathbf{T}^{n+1}
-
\Delta t^{-1}\mathbf{M}\mathbf{T}^{n}
\right),
\label{eq:adjoint_lagrangian}
\end{align}
where $N$ is the number of time steps and $\boldsymbol{\lambda}^{n}$ is
the discrete adjoint variable associated with the forward step from
$t^{n}$ to $t^{n+1}$. Requiring stationarity of $\mathcal{L}$ with respect
to each $\mathbf{T}^{n+1}$ yields the adjoint equations, which are solved
backward in time,
\begin{align}
\mathbf{A}^{\top}\boldsymbol{\lambda}^{n}
=
\Delta t^{-1}\mathbf{M}\boldsymbol{\lambda}^{n+1}
+
\frac{\partial\mathcal{J}}{\partial\mathbf{T}^{n+1}},
\qquad
n=N-1,\ldots,0,
\qquad
\boldsymbol{\lambda}^{N}=\mathbf{0},
\label{eq:adjoint_system}
\end{align}
with the adjoint source given by the observation mismatch,
\begin{align}
\frac{\partial\mathcal{J}}{\partial\mathbf{T}^{k}}
=
\begin{cases}
\mathbf{H}^{\top}\left(\mathbf{H}\mathbf{T}^{k}-\mathbf{T}^{\mathrm{obs}}_{k}\right),
& k\in\mathcal{K}_{\mathrm{obs}},\\[2pt]
\mathbf{0}, & \text{otherwise}.
\end{cases}
\label{eq:adjoint_source}
\end{align}
Since $\mathbf{A}$ depends on the boundary parameters only through
$\mathbf{B}_{fc}$, differentiating $\mathcal{L}$ with respect to an
individual boundary parameter gives
\begin{align}
\frac{d\mathcal{J}}{db_j}
=
-\sum_n
\int_{\Gamma_c}
\phi_j(x)T^{n+1}\lambda^n
\,\dG
+
\frac{\partial\mathcal{J}_{\mathrm{reg}}}
{\partial b_j},
\label{eq:gradient}
\end{align}
where $T^{n+1}$ and $\lambda^{n}$ denote the finite-element functions with
nodal values $\mathbf{T}^{n+1}$ and $\boldsymbol{\lambda}^{n}$,
respectively. Thus, the
observation mismatch is propagated backward through the adjoint equations
to obtain the gradient of the inverse objective with respect to the
distributed boundary parameters, without requiring a separate sensitivity
solve for each parameter.

The boundary control parameters are obtained by solving the
bound-constrained optimization problem
\begin{equation}
\min_{\mathbf{b}\in\mathcal{B}} \mathcal{J}(\mathbf{b}),
\label{eq:opt_problem}
\end{equation}
where $\mathbf{b}=[b_1,\ldots,b_{N_p}]^T$ contains the $N_p$ boundary
control parameters and $\mathcal{B}$ denotes their prescribed admissible
range. The optimization is performed using the L-BFGS-B algorithm with the
gradient obtained from the discrete adjoint equations.

\begin{figure}[htbp]
\centering
\resizebox{\ifdim\width>\linewidth\linewidth\else\width\fi}{!}{%
\includegraphics{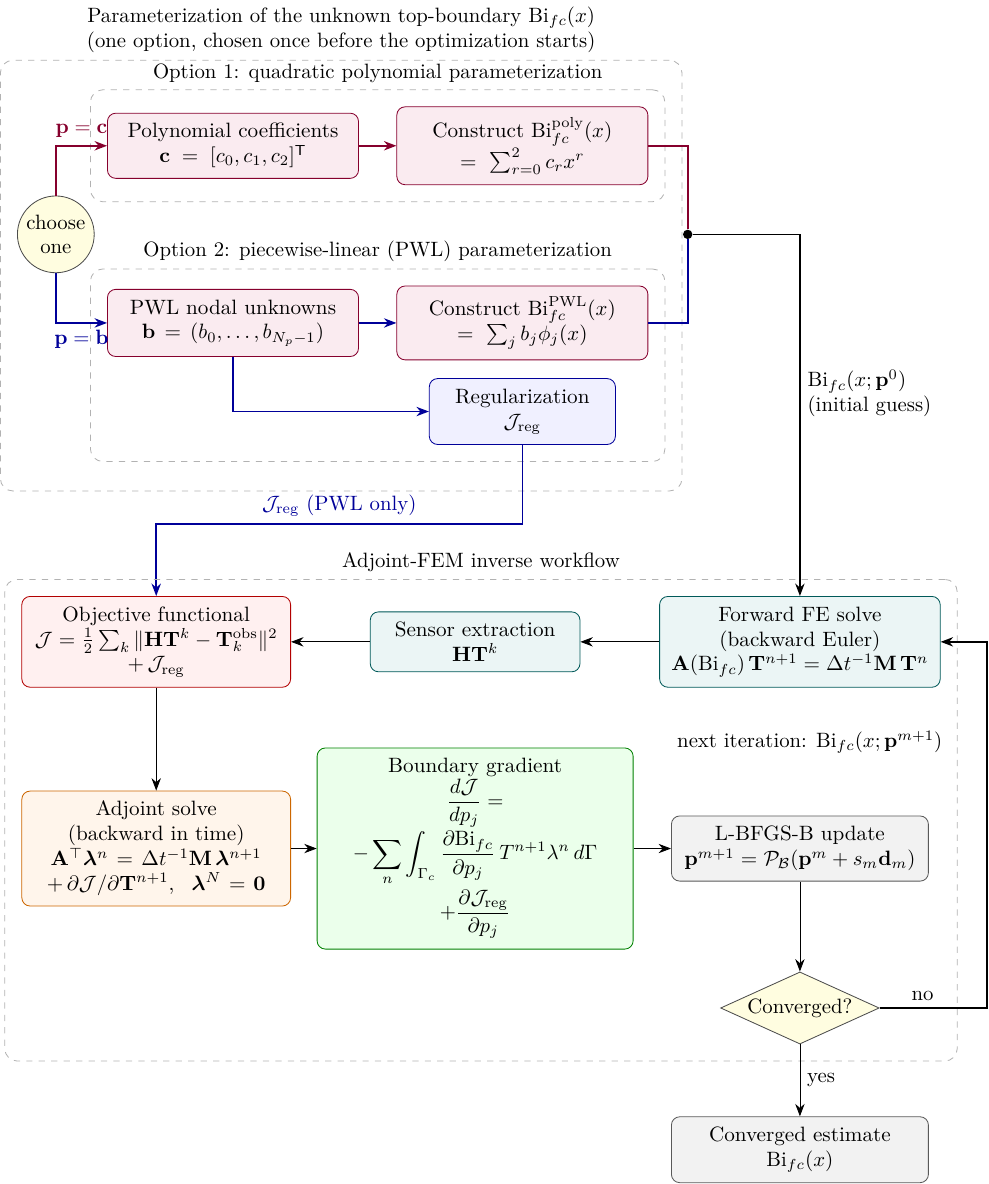}%
}
\caption{Workflow of the Adjoint-FEM inverse framework. The unknown top-boundary
Biot profile is represented by one of two alternative parameterizations,
selected once before the reconstruction: a low-dimensional quadratic polynomial
(\Cref{eq:polynomial_bi}) or a generalized piecewise-linear (PWL) nodal
parameterization (\Cref{eq:nodal_bi_fem}). Only one parameterization is active in a
given reconstruction, and $\mathbf{p}=\mathbf{c}$ or $\mathbf{p}=\mathbf{b}$
denotes the corresponding parameter vector. Each iteration performs a
backward-Euler forward finite-element solve, extracts sensor predictions,
evaluates the objective in \Cref{eq:cost_functional}, solves the discrete
adjoint problem in \Cref{eq:adjoint_system,eq:adjoint_source} backward in
time, and computes the top-boundary gradient in \Cref{eq:gradient}.
The boundary sensitivity $\partial\Bi_{fc}/\partial p_j$ is $x^j$ for the
quadratic coefficients ($j=0,1,2$) and $\phi_j(x)$ for PWL nodal values.
L-BFGS-B then updates $\mathbf{p}$, shown schematically with the
limited-memory quasi-Newton search direction $\mathbf{d}_m$, step length
$s_m$, and projection $\mathcal{P}_{\mathcal{B}}$ onto the admissible bounds
$\mathcal{B}$. The iterations terminate once the objective and profile
stopping criteria in \Cref{subsec:app_optimization} are satisfied. For PWL
reconstructions, the $H^1$-seminorm regularization in
\Cref{eq:adjoint_regularization} is added to the objective to suppress
unsupported spatial oscillations in the high-dimensional inverse profile.}
\label{fig:adjoint_architecture}
\end{figure}

The resulting forward--adjoint optimization procedure is summarized in
\Cref{fig:adjoint_architecture}. Starting from an initial boundary control, the
forward problem is solved, the resulting temperature field is compared
with the observations, and the discrete adjoint problem is then solved
backward in time to obtain the gradient with respect to the boundary
control. The control is subsequently updated using the bound-constrained
optimizer, and the procedure is repeated until the prescribed convergence
criteria are satisfied.

%

\section{Numerical setup}
\label{sec:setup}

\subsection{Data Generation and Measurement Configuration}
\label{subsec:data_generation}

For the jet-impingement cases, three-dimensional CFD--CHT simulations are
used to generate the temperature data used as observations by the inverse
methods. The temperature field is sampled at the prescribed sensor
locations and observation times, after which synthetic measurement noise is
introduced to emulate uncertainty in temperature measurements. The
corresponding CFD--CHT heat-transfer coefficient distributions are retained
as independent high-fidelity reference estimates for evaluating the
reconstructed boundary fields
\citep{hazra_pinnbased_2026a}.

For the controlled source problem, temperature observations are generated
by solving the nondimensional transient heat-conduction model in ~\Cref{eq:governing_nondim} with a prescribed Gaussian top-boundary profile,
\begin{align}
\Bi_{fc}^{G}(x)
=
\Bi_{\min}
+
\left(\Bi_{\max}-\Bi_{\min}\right)
\exp\left[
-\frac{(x-x_c)^2}{2\sigma^2}
\right],
\label{eq:gaussian_source_profile}
\end{align}
where $\Bi_{\min}$ and $\Bi_{\max}$ denote the minimum and maximum
boundary coefficients, $x_c$ is the location of the Gaussian center,
and $\sigma$ controls its spatial width. The resulting temperature field is sampled at the selected sensor locations
and observation times. The source observations are generated using the same
two-dimensional forward model employed in the inverse problem, providing a
controlled source problem for the transfer-learning experiments.

The CHT-derived target cases considered in this study correspond to Reynolds's number \numlist{5000;10000;15000}, consistent
with the Reynolds-number cases investigated in our previous study
\citep{hazra_pinnbased_2026a}.
Following that
study, the jet Reynolds number is defined based on the nozzle diameter
and jet-exit velocity as
\begin{align}
	Re = \frac{U_j D_n}{\nu},
	\label{eq:reynolds_number}
\end{align}
where $U_j$ is the jet-exit velocity, $D_n=\SI{6}{\milli\meter}$ is the nozzle diameter,
and $\nu=\SI{1.516e-5}{\meter^2\per\second}$ is the kinematic viscosity of air. The corresponding jet-exit velocities are
\SI{12.63}{\meter\per\second}, \SI{25.26}{\meter\per\second}, and
\SI{37.89}{\meter\per\second} for $Re=5000$, $10000$, and $15000$,
respectively.
The same Reynolds-number cases and
CFD--CHT data-generation framework are retained here to provide
continuity with the previously established data set while allowing the
present study to focus on generalized boundary representations and
transfer-learning strategies.

The numerical experiments are organized into three groups:
(i) boundary-representation comparison, (ii) ransfer between noise levels at the same Reynolds number, and (iii) transfer from a controlled Gaussian-profile problem to CFD–CHT-derived target problems at different Reynolds numbers. The corresponding
source and target problems, observation intervals, noise levels,
and PINN boundary representations are summarized in
\Cref{tab:case_matrix}. The boundary-representation comparison considers
the quadratic, PWL, and NN representations for the
$\mathrm{Re}=5000$ CHT-derived case at two temporal observation
resolutions. The fixed-Reynolds-number transfer experiments use a
noise-free CHT-derived case at $\mathrm{Re}=5000$ as the source and noisy
observations of the same physical case as the targets. The Gaussian-to-CHT transfer experiments use the controlled Gaussian problem as the
source and CHT-derived cases at $\mathrm{Re}=5000,\ 10000,\ \text{and }15000$ as the
targets with \SI{5}{\percent} additive Gaussian noise applied to the target
temperature observations to represent a moderate level of experimental measurement uncertainty.

\begin{table}[htbp]
	\centering
	\caption{Summary of inverse-problem cases considered in the numerical
		study. Each target case is also reconstructed with Adjoint-FEM using
		the PWL representation, and additionally the quadratic representation
		in the boundary-representation comparison; the NN representation is
		used only within the PINN framework.}
	\label{tab:case_matrix}
	\renewcommand{\arraystretch}{1.2}
	\small
	\setlength{\tabcolsep}{4pt}
	\begin{tabular}{@{}>{\raggedright\arraybackslash}p{3.6cm}
			>{\raggedright\arraybackslash}p{2.6cm}
			>{\raggedright\arraybackslash}p{2.8cm}
			>{\raggedright\arraybackslash}p{1.2cm}
			>{\raggedright\arraybackslash}p{1.3cm}
			>{\raggedright\arraybackslash}p{2.6cm}@{}}
		\toprule
		\textbf{Comparison}
		& \textbf{Source problem}
		& \textbf{Target problem}
		& $\Delta t_{\mathrm{obs}}$
		& \textbf{Noise}
		& \textbf{PINN representations} \\
		\midrule
		Boundary-representation comparison
		& --
		& CHT, $\mathrm{Re}=5000$
		& \SI{1}{\second}, \SI{4}{\second}
		& \SI{5}{\percent}
		& Quadratic, PWL, NN \\
		\addlinespace[4pt]
		Fixed-Reynolds-number transfer
		& CHT, $\mathrm{Re}=5000$, \SI{0}{\percent}
		& CHT, $\mathrm{Re}=5000$
		& \SI{1}{\second}
		& \SI{5}{\percent}, \SI{10}{\percent}
		& PWL, NN \\
		\addlinespace[4pt]
		Gaussian-to-CHT transfer
		& Gaussian, \SI{0}{\percent}
		& CHT, $\mathrm{Re}=5000,\ 10000,\ 15000$
		& \SI{1}{\second}
		& \SI{5}{\percent}
		& PWL, NN \\
		\bottomrule
	\end{tabular}
\end{table}

Both inverse methods use the same four sensor locations within the
\SI{60}{\milli\meter} $\times$ \SI{15}{\milli\meter} two-dimensional
domain. The sensors are located \SI{0.9375}{\milli\meter} beneath the
impingement surface at
\begin{align}
	(x,y)
	&=
	(3.75,14.0625),\,
	(18.75,14.0625),\,
	(33.75,14.0625),\,
	(48.75,14.0625)\,\si{\milli\meter}.
	\label{eq:sensor_locations}
\end{align}

The forward simulation spans
$\SIrange{0}{300}{\second}$, with temperature snapshots available at
\SI{0.25}{\second} intervals. Two observation schedules are extracted from
this common temporal grid:
\begin{align}
	\Delta t_{\mathrm{obs}}
	&= \SI{1}{\second},
	&
	t
	&=
	\{0,1,2,\ldots,300\}\,\si{\second},
	\label{eq:measurement_res1}
	\\
	\Delta t_{\mathrm{obs}}
	&= \SI{4}{\second},
	&
	t
	&=
	\{0,4,8,\ldots,300\}\,\si{\second}.
	\label{eq:measurement_res4}
\end{align}
The \SI{1}{\second} and \SI{4}{\second} observation intervals are
selected from the temporal sampling-rate range of
$0.25$--$4\,\si{\hertz}$ investigated in our previous study
\citep{hazra_pinnbased_2026a}. These two intervals are used in the
boundary-representation comparison to examine the effect of temporal
data availability on the reconstruction of the spatially varying HTC
while keeping the underlying physical case unchanged. The
transfer-learning experiments use the \SI{1}{\second} observation
interval throughout so that temporal observation resolution does not
introduce an additional variation in the comparison.

The observation interval denotes the spacing between retained temperature
observations and is independent of the spatial discretization used to
represent the unknown boundary condition. In particular, it does not
change the CFD--CHT boundary profile or the number of PWL boundary
parameters. The PWL representation uses $N_p=200$ boundary parameters
throughout the inverse-estimation and transfer-learning calculations; this
resolution is a fixed numerical choice and is not treated as an optimization
variable. The CFD--CHT reference HTC, however, is available only at 60
locations along the impingement surface. For quantitative comparison, the
200-node PWL reconstructions are therefore linearly interpolated onto these
60 reference locations, while the quadratic and NN-estimator profiles are
evaluated there directly. In the profile figures, PWL curves are shown on
their native 200-point grids and the quadratic and NN-estimator curves at the
60 reference points.

Synthetic Gaussian noise is added after temporal downsampling. Let
$T_{ij}$ denote the nondimensional temperature at observation time $t_i$
and sensor $j$. The instantaneous mean and standard deviation across the
sensor array are defined as
\begin{align}
	\bar{T}_i
	&=
	\frac{1}{N_s}
	\sum_{j=1}^{N_s} T_{ij},
	\\
	\sigma_i
	&=
	\left[
	\frac{1}{N_s}
	\sum_{j=1}^{N_s}
	\left(T_{ij}-\bar{T}_i\right)^2
	\right]^{1/2}.
\end{align}
The noisy observations are then generated according to
\begin{equation}
	T^{\mathrm{meas}}_{ij}
	=
	T_{ij}
	+
	\eta\,\sigma_i\,\epsilon_{ij},
	\qquad
	\epsilon_{ij}\sim\mathcal{N}(0,1),
	\label{eq:noise_model}
\end{equation}
where $\eta$ denotes the prescribed dimensionless noise level. The
principal comparisons use $\eta=\num{0}$, $\num{0.05}$, and
$\num{0.10}$, corresponding to noise-free, \SI{5}{\percent}, and
\SI{10}{\percent} observations, respectively. Because $\sigma_i$ is
evaluated independently at each observation time, the magnitude of the
perturbation varies with the instantaneous spatial variation of the sensor
temperatures.

For each paired comparison, the same noisy temperature observations are
supplied to the PINN and Adjoint-FEM. The PINN observation array retains
the $t=0$ data, whereas the Adjoint-FEM data-misfit objective begins at
the first positive observation time. Since all four nondimensional sensor
temperatures equal unity at $t=0$, the additional PINN data row is an
unperturbed representation of the prescribed initial condition and does
not provide independent information about the boundary coefficient.

\subsection{PINN and Transfer-Learning Configuration}
\label{subsec:pinn_setup}

The temperature field in PINN formulations is represented by fully connected neural networks with
$\tanh$ activations and trained using the Adam optimizer. Fixed Sobol
sequences are used to generate the interior, boundary, and
initial-condition collocation points.

The production configurations are selected through an offline
hyperparameter-selection procedure using the tree-structured Parzen
estimator (TPE) ~\citep{bergstra_algorithms_2011} implemented in Optuna ~\citep{akiba_optuna_2019}. The search considers the network
depth and width, collocation-point allocation, learning rate, network-weight
regularization, and loss-weight refinement parameters. Candidate models are
trained in 1000-iteration chunks, with median pruning activated after five
protected chunks. The selected configurations are subsequently used for
the production reconstructions. Hyperparameter-search cost is treated as
an offline model-selection cost and is excluded from the reported
production-training times. The selected production configurations for the controlled Gaussian source
and CHT-derived target problems are summarized in
\Cref{tab:pinn_config}.


\begin{table}[!htbp]
	\centering
	\caption{Production PINN configurations used for the Gaussian source and
		CHT-derived inverse problems.}
	\label{tab:pinn_config}
	\renewcommand{\arraystretch}{1.15}
	\small
	\setlength{\tabcolsep}{4pt}
	\begin{tabular}{@{}lcccc@{}}
		\toprule
		\textbf{Parameter}
		& \textbf{Gaussian source}
		& $\boldsymbol{\mathrm{Re}=5000}$
		& $\boldsymbol{\mathrm{Re}=10000}$
		& $\boldsymbol{\mathrm{Re}=15000}$ \\
		\midrule
		Hidden layers
		& 5 & 3 & 3 & 5 \\
		Neurons per layer
		& 96 & 96 & 96 & 64 \\
		Activation
		& tanh & tanh & tanh & tanh \\
		Optimizer
		& Adam & Adam & Adam & Adam \\
		Learning rate
		& \num{6.698e-4}
		& \num{5.723e-4}
		& \num{2.476e-4}
		& \num{2.545e-4} \\
		Maximum iterations
		& 100000 & 50000 & 50000 & 100000 \\
		Domain points
		& 2542 & 6242 & 13545 & 8070 \\
		Boundary points
		& 2198 & 4175 & 3999 & 3600 \\
		Initial-condition points
		& 571 & 2796 & 4093 & 659 \\
		Test points
		& 3000 & 1000 & 1000 & 1000 \\
		\bottomrule
	\end{tabular}
\end{table}
The NN-estimator boundary branch has the architecture $[1,32,32,1]$,
consisting of two hidden layers with 32 $\tanh$ units each and a linear
scalar output. This branch contains 1153 trainable weights and biases. No
explicit boundary-profile regularization or output clipping is applied to
the NN-estimator branch. The right-boundary coefficient $\Bi_{fr}$ is fixed
in all reported cases.

For the PWL boundary representation, explicit boundary-profile smoothing is disabled for all reported scratch-trained and transferred PWL target models. The only exception is the noise-free \(\mathrm{Re}=5000\). PWL source used in the fixed-Reynolds-number transfer pathway, for which an \(\ell_1\) smoothing penalty with coefficient \(\num{3.0e-04}\) is employed. The corresponding smoothing term is not retained during optimization of the transferred target models so that the target reconstruction is determined solely by the target observations and governing equations, while the source regularization serves only to provide a stable initialization.

The loss-weight initialization defined in \Cref{subsec:pinn}
is implemented using a 500-iteration warm-up with all loss weights
initially set to unity. The loss magnitudes are floored at \num{1.0e-08}
before forming the relative weights, with base weights constrained to
$[0.1,100]$. During offline hyperparameter selection, the base weights
are refined using multiplicative factors
$\gamma_i\sim\operatorname{LogU}(0.1,10)$, with the resulting weights
constrained to $[0.1,1000]$. If the warm-up procedure fails or produces
non-finite loss values, the corresponding weights are sampled directly
from $\operatorname{LogU}(1,50)$. The selected loss weights are then
fixed for the corresponding production run.

All reported production PINNs employ early stopping after a minimum of
5000 iterations. Since the inverse objective concerns the reconstruction
of the unknown boundary quantity, the stopping criterion monitors
stabilization of the estimated boundary representation rather than
relying solely on stagnation of the total PINN loss.

Two transfer-learning pathways are considered, following the methodology
described in \Cref{subsubsec:transfer_learning}. In the fixed-Reynolds-number
pathway, a noise-free $\mathrm{Re}=5000$ source model is transferred to
$\mathrm{Re}=5000$ target problems with 5\% and 10\% observation noise.
This pathway retains the same physical problem while changing the
observation quality. In the Gaussian-to-CHT transfer, the controlled
Gaussian source models are transferred to CHT-derived target problems at
$\mathrm{Re}=5000$, $10000$, and $15000$ with 5\% observation noise. This
pathway changes both the target boundary-profile characteristics and the
data-generation setting.

For the fixed-Reynolds-number pathway, the source model is trained using the
noise-free observations and subsequently adapted to the \SI{5}{\percent} and
\SI{10}{\percent} measurement-noise target problems. The boundary parameterization and
NN architecture, where applicable, are kept unchanged between the source
and target models. During target adaptation, only the final hidden layer and
output layer of the temperature network are updated, while the preceding
temperature-network layers are kept fixed. The target boundary variables
remain trainable.

For the Gaussian-to-CHT transfer pathway, the source model is trained for the
controlled Gaussian inverse problem and subsequently adapted to the CHT
targets at $\mathrm{Re}=$\numlist{5000;10000;15000}. The target temperature
networks are initialized using pretrained Gaussian-source weights. The
source and target temperature networks use compatible architectures for
weight transfer. For NN-estimator-to-NN-estimator transfer, the auxiliary
boundary networks also use matching architectures. Four of the five hidden layers of the Gaussian source
temperature network are frozen during target adaptation, while the final
hidden layer, output layer, and target boundary variables remain trainable.
For NN-parameterization-based transfer, the source boundary-network parameters are also
used to initialize the corresponding target boundary branch and remain
trainable during target optimization.

For each target problem, the transferred model is compared with a
scratch-trained PINN using the same target observations, noise realization,
and boundary representation. Thus, both models solve the same target inverse problem, differing only in their initialization and, for the transferred model, the prescribed layer-freezing strategy.

\subsection{Discrete Adjoint-FEM Implementation Details}
\label{subsec:adjoint_numerical_setup}

The forward and adjoint equations are discretized using continuous
piecewise-linear finite elements on a structured mesh and integrated in time
using the backward Euler scheme. The generalized PWL control uses 200
top-boundary nodes, corresponding to a $199\times16$ forward finite-element
mesh, while the low-dimensional quadratic representation uses a
$64\times16$ forward mesh. All reconstructions are initialized with
\begin{align}
	\Bi_{fc,0}(x)
	&=1-x.
	\label{eq:adjoint_initial_profile}
\end{align}

For the generalized PWL reconstructions, the boundary coefficient is
constrained pointwise according to
\begin{align}
	0\leq\Bi_{fc}(x)
	&\leq2.
	\label{eq:adjoint_biot_bounds}
\end{align}
This admissible range is prescribed a priori and is used for all Reynolds
numbers, observation resolutions, and noise levels considered in the
generalized PWL cases. The three coefficients of the quadratic
representation are subject to their corresponding prescribed bounds.

The generalized PWL reconstructions employ the $H^1$-seminorm regularization
defined in \Cref{eq:adjoint_regularization}.
The regularization parameter is examined using a seven-point logarithmically
spaced grid over the study-specific interval $[10^{-3},10^{-3/11}]$,
\begin{align}
	\mu_{\mathrm{reg}}
	\in
	\{
	&\num{0.001},
	\num{0.00284804},
	\num{0.00811131},
	\num{0.02310130},
	\num{0.06579332},
	\num{0.18738174},
	\num{0.53366992}
	\}.
	\label{eq:seven_point_grid}
\end{align}
The logarithmic construction of this grid is detailed in
\Cref{subsec:app_pwl_regularization}.
The grid provides a common set of candidate values for the L-curve
diagnostics while limiting the number of offline inverse solves. The
resulting L-curves are used to assess the transition between data
fidelity and profile regularization. For the controlled comparison,
the common sampled value
$\mu_{\mathrm{reg}}=\num{0.02310130}$ is adopted for all reported
production PWL reconstructions. The quadratic representation does not
employ explicit profile regularization. Regularization-selection
calculations are treated as offline cost and are excluded from the
production computational-time comparison.

The bound-constrained optimization is performed using L-BFGS-B. For the
quadratic reconstructions, the L-BFGS-B convergence criteria are applied
using the prescribed relative-objective and projected-gradient tolerances.
For the generalized PWL reconstructions, these criteria are supplemented by
a profile-stabilization criterion based on the change in the objective and
the reconstructed PWL control over successive iterations. The
profile-stabilization criterion is evaluated over a three-iteration window,
with convergence requiring two consecutive successful checks after a
minimum of 10 iterations. The numerical tolerances and iteration limits are summarized in Appendix~\ref{app:adjoint_fem}

For the quadratic reconstructions, the relative-objective tolerance is set to
$\tau_f=\num{1.0e-14}$ or $\num{1.0e-12}$, depending on the observation
configuration, while the projected-gradient tolerance is fixed at
$\tau_g=\num{1.0e-8}$. The corresponding maximum iteration limits are
selected for each production configuration. For the generalized PWL
reconstructions, $\tau_f=\num{1.0e-14}$ and
$\tau_g=\num{1.0e-8}$ are used together with the additional profile-aware
stopping criterion described in \Cref{subsec:adjoint}. Specifically, both
the objective and the PWL control are required to change by less than
\SI{0.5}{\percent} over a three-iteration window, with the condition
satisfied on two consecutive checks and subject to a minimum of 10
iterations. 

\subsection{Evaluation Metrics and Computational Environment}
\label{subsec:performance_metrics}

The reconstructed boundary condition is evaluated in terms of the dimensional
heat-transfer coefficient $h$. For the CHT-derived cases, each estimated
profile is evaluated on the reference grid defined in
\Cref{subsec:data_generation} before computing
the error metrics. The relative spatial $L^2$ error is defined as
\begin{align}
	\varepsilon_h
	=
	100\,
	\frac{\left\|\mathbf{h}^{\mathrm{est}}
		-\mathbf{h}^{\mathrm{ref}}\right\|_2}
	{\left\|\mathbf{h}^{\mathrm{ref}}\right\|_2},
	\label{eq:htc_relative_error}
\end{align}
and the dimensional root-mean-square error (RMSE) is
\begin{align}
	\mathrm{RMSE}_h
	=
	\left[
	\frac{1}{N_h}
	\sum_{i=1}^{N_h}
	\left(
	h_i^{\mathrm{est}}-h_i^{\mathrm{ref}}
	\right)^2
	\right]^{1/2}.
	\label{eq:htc_rmse}
\end{align}
The RMSE is reported in
\si{\watt\per\meter\squared\per\kelvin}.

The consistency of the reconstructed temperature field with the
measurements is assessed using the observation-normalized $L^2$
discrepancy over the common sensor--time data set. This quantity is used
to verify agreement with the observations and is considered together
with the HTC reconstruction error rather than as an independent measure
of the quality of the recovered boundary profile.

All computations were performed on a single, consistent computing setup featuring an Intel\reg Xeon\reg Silver 4516Y+ 48-core CPU (2.0 GHz base clock, up to 3.7 GHz boost), 64 GB of DDR5 memory, and an NVIDIA RTX A5000 GPU with 24 GB of dedicated VRAM running CUDA 12.8. The machine used Ubuntu 24.04 LTS and Python 3.9.18, with CUDA 12.8 and TensorFlow 2.20.0 for PINN-related computations. PINNs were implemented and trained with the DeepXDE framework~\citep{lu_deepxde_2021}, whereas adjoint-FEM computations were conducted using FEniCSx 0.10.1~\citep{baratta_dolfinx_2023}. CPU thread counts and the number of GPU devices were held constant for all production computations. Each production run was performed in isolation, with no other heavy workloads running at the same time, in order to reduce variability in recorded wall-clock times. 

Wall-clock time was measured from the start of the production
optimization or training procedure to its termination according to the
prescribed stopping criterion. For scratch-trained PINNs, the reported
time therefore corresponds to the complete target-model training
procedure. For transferred PINNs, timing begins after transfer
initialization and includes only the subsequent target-model training;
source-model training is treated as an offline cost and is excluded from
the reported target-estimation time. For Adjoint-FEM, the reported time
covers the final PWL inverse optimization and excludes the independent
quadratic reconstruction and the offline regularization-parameter
selection. Hyperparameter searches, including the Optuna search used for
PINN configuration and loss-weight selection, are likewise excluded from
the production timings.

For each transferred target, the computational reduction relative to the
corresponding scratch-trained PINN is quantified by
\begin{align}
	S_t
	=
	100
	\frac{
		t_{\mathrm{scratch}}-t_{\mathrm{TL}}
	}{
		t_{\mathrm{scratch}}
	},
	\label{eq:transfer_speedup}
\end{align}
where positive $S_t$ denotes a reduction in target-estimation time.
Computational savings are considered together with the corresponding
reconstruction errors when assessing the effect of transfer learning.

%

\section{Results and discussion}
\label{sec:results}

The results assess the generalized boundary representations and
transfer-learning strategies introduced in \Cref{sec:methodology} for
inverse estimation of the spatially varying convective heat transfer
coefficient.  The analysis is organized
around two questions. The analysis addresses two questions: whether generalized
boundary representations can retain reconstruction accuracy relative to
the quadratic representation, and whether transfer learning can reduce
the cost of repeated inverse estimation relative to training a PINN from
scratch.

\subsection{From parametric to generalized HTC reconstruction}
\label{subsec:validation}
The first set of results examines the quadratic, PWL, and NN-estimator
representations for the $\mathrm{Re}=5000$ CHT target. The reconstructed
HTC profiles are evaluated against the corresponding high-fidelity
CFD--CHT reference, with two temporal observation intervals considered to
examine the effect of reduced temporal data availability. The quadratic
and PWL Adjoint-FEM reconstructions are included as independent
deterministic comparisons. The same representation comparison is then performed for the \(\mathrm{Re}=15000\) target to examine whether the representation-dependent reconstruction behavior persists for a different CHT-derived HTC profile.

\subsubsection{Effect of representation and temporal observation interval}
\label{subsubsec:representation_temporal}

\begin{figure}[!hbtp]
\centering
\includegraphics{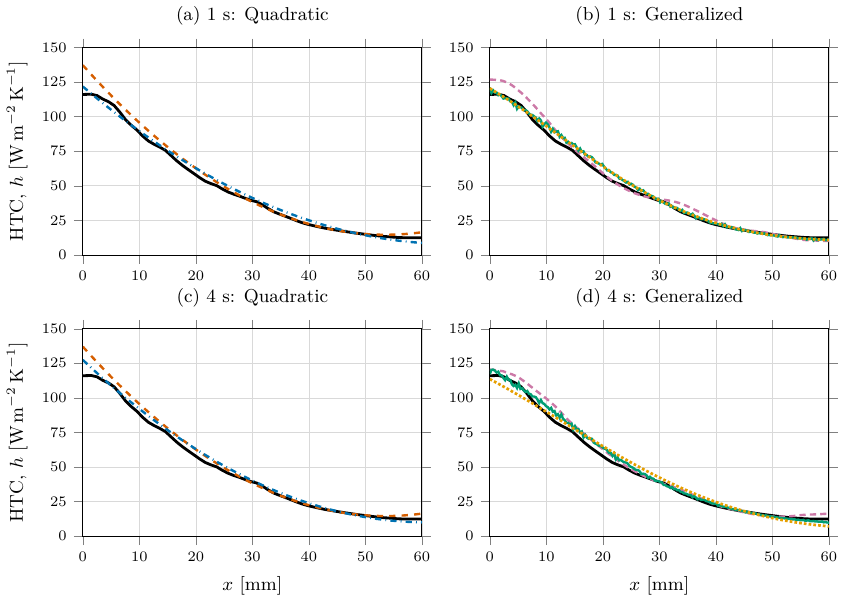}

\vspace{0.4em}
\includegraphics{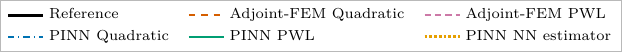}
\caption{Reference and reconstructed HTC profiles for the representative
$\mathrm{Re}=5000$ case with 5\% measurement noise. The quadratic panels show
the Adjoint-FEM and PINN estimates; the generalized panels show the
Adjoint-FEM PWL, PINN PWL, and PINN NN-estimator reconstructions. Results are
shown separately for the 1-s and 4-s sampling intervals.}
\label{fig:table7_htc_profiles}
\end{figure}

The representative $\mathrm{Re}=5000$ CHT target is considered with
$5\%$ additive Gaussian noise and two temporal observation intervals,
$\Delta t_{\mathrm{obs}}=1$ and $\SI{4}{\second}$. At each observation
resolution, the quadratic, PWL, and NN-estimator PINNs are compared with
the corresponding quadratic and generalized PWL Adjoint-FEM
reconstructions. The reconstructed HTC profiles are shown in
\Cref{fig:table7_htc_profiles}.

At the $\SI{1}{\second}$ observation interval, all three PINN
representations reproduce the principal monotonic decrease of the HTC
profile. The PWL and NN-estimator profiles remain smooth
over the impingement surface, with no pronounced high-frequency
oscillations. The
reconstructed profiles are visually close to the reference profile.
The corresponding quantitative results are summarized in
\Cref{tab:representative_generalized}. At the $\SI{1}{\second}$
observation interval, the relative profile errors of the quadratic, PWL,
and NN-estimator PINNs are \SI{5.08}{\percent}, \SI{5.31}{\percent}, and \SI{5.19}{\percent},
respectively. The corresponding quadratic and generalized PWL Adjoint-FEM reconstructions give relative errors of \SI{8.68}{\percent} and \SI{8.75}{\percent}, respectively.
Thus, for the $\mathrm{Re}=5000$ target, increasing the flexibility of the boundary representation does not provide a noticeable reduction in reconstruction error for either inverse method.
Increasing the observation interval from 1 to \SI{4}{\second} produces only modest changes in reconstruction error for the quadratic and PWL representations, whereas the NN-estimator shows a more noticeable increase in error. The quadratic PINN, PWL PINN, and both Adjoint-FEM representations therefore show limited sensitivity to the reduction in temporal observation density over the two intervals considered, whereas the NN-estimator is more sensitive. The effect of temporal data density depends on the
boundary representation and inverse formulation rather than on the
number of observations alone.

\def\RepresentativeTableCaption{Representation comparison for the
$\mathrm{Re}=5000$ CHT profile under 5\% measurement noise. The quadratic
and generalized PWL Adjoint-FEM formulations are evaluated alongside
the corresponding PINN representations.}
\def\RepresentativeTableLabel{tab:representative_generalized}
\def\RepresentativeTableRows{
$\mathrm{Re}=5000$ & 1 s & 5\% & Quadratic
  & Adjoint-FEM Quadratic & 5.06 & 8.68 & 228.5 \\
 &  &  &  & PINN Quadratic & 2.96 & 5.08 & 105.7 \\
\cmidrule(l){4-8}
&  &  & Generalized
  & Adjoint-FEM PWL & 5.10 & 8.75 & 537.6 \\
 &  &  &  & PINN PWL & 3.09 & 5.31 & 315.5 \\
 &  &  &  & PINN NN estimator & 3.02 & 5.19 & 356.4 \\
\midrule
$\mathrm{Re}=5000$ & 4 s & 5\% & Quadratic
  & Adjoint-FEM Quadratic & 5.03 & 8.63 & 151.1 \\
 &  &  &  & PINN Quadratic & 3.11 & 5.33 & 228.1 \\
\cmidrule(l){4-8}
&  &  & Generalized
  & Adjoint-FEM PWL & 4.57 & 7.83 & 592.9 \\
 &  &  &  & PINN PWL & 3.03 & 5.20 & 252.6 \\
 &  &  &  & PINN NN estimator & 4.35 & 7.47 & 140.9 \\
}
\begin{table}[H]
\centering
\caption{\RepresentativeTableCaption}
\label{\RepresentativeTableLabel}
\renewcommand{\arraystretch}{1.2}
\small
\setlength{\tabcolsep}{5pt}
\begin{tabular}{@{}c c c l l
                S[table-format=2.2]
                S[table-format=2.2]
                S[table-format=3.1]@{}}
\toprule
\textbf{Target} &
$\boldsymbol{\Delta t_{\mathrm{obs}}}$ &
\textbf{Noise} &
\textbf{Estimator} &
\textbf{Solver} &
{\shortstack{$\boldsymbol{\mathrm{RMSE}_{h}}$\\{\footnotesize(\si{\watt\per\meter\squared\per\kelvin})}}} &
{$\boldsymbol{\varepsilon_{h}}$ \textbf{(\%)}} &
{\textbf{Time (s)}} \\
\midrule
\RepresentativeTableRows
\bottomrule
\end{tabular}
\end{table}

The computational behavior also differs between the two inverse
approaches and with the temporal observation density. For the
Adjoint-FEM formulation, replacing the quadratic boundary
representation with the 200-parameter PWL representation substantially
increases the computational cost at both observation intervals,
reflecting the increased dimensionality of the boundary control and the
associated forward--adjoint optimization procedure. In contrast, the
PINN computational cost does not increase monotonically with the
flexibility of the boundary representation or with the number of
retained observations. At the $\SI{1}{\second}$ observation interval,
the PWL and NN-estimator require more computational time than the
quadratic PINN, whereas at $\SI{4}{\second}$ the NN-estimator requires
substantially less time than both the quadratic and PWL PINNs. Thus,
reducing the temporal observation density does not translate directly
into a proportional reduction in the total computational cost of
inverse estimation, and the effect differs between the PINN and
Adjoint-FEM formulations.

\subsubsection{Representation comparison for the higher-Reynolds-number target}
\label{subsec:re15000_representation}

\begin{figure}[H]
\centering
\includegraphics{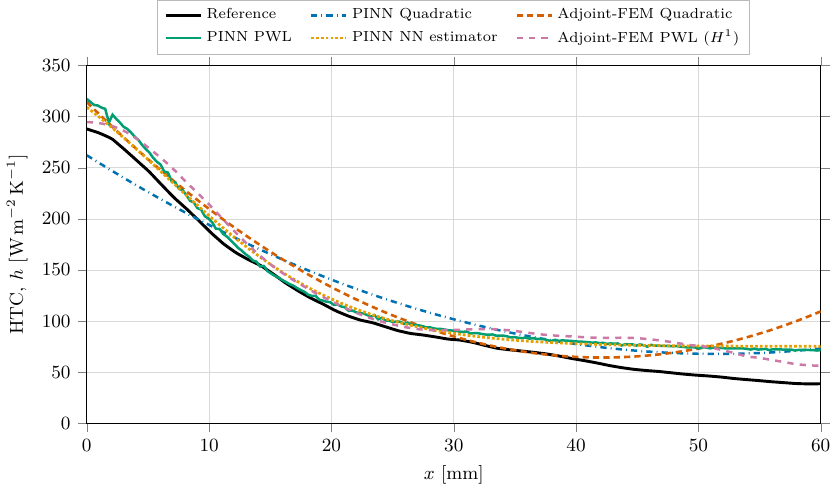}
\caption{Reference and reconstructed HTC profiles for the
$\mathrm{Re}=15000$ CHT target with 5\% measurement noise and the 1-s
sampling interval. The three scratch-trained PINN boundary representations are
compared with the quadratic and generalized PWL Adjoint-FEM reconstructions.}
\label{fig:re15000_representation_comparison}
\end{figure}

The representation comparison is repeated for the
$\mathrm{Re}=15000$ CHT target using $5\%$ measurement noise and the
$\SI{1}{\second}$ observation interval. The quadratic, PWL, and
NN-estimator PINNs are compared with the corresponding quadratic and
generalized PWL Adjoint-FEM reconstructions. The reconstructed profiles
are shown in \Cref{fig:re15000_representation_comparison}, and the
corresponding errors and computational times are summarized in
\Cref{tab:re15000_representation}.

\def\RepresentativeTableCaption{Representation comparison for the
$\mathrm{Re}=15000$ CHT profile under 5\% measurement noise using the 1-s
observation interval.}
\def\RepresentativeTableLabel{tab:re15000_representation}
\def\RepresentativeTableRows{
$\mathrm{Re}=15000$ & 1 s & 5\% & Quadratic
  & Adjoint-FEM Quadratic & 24.99 & 18.96 & 141.0 \\
 &  &  &  & PINN Quadratic & 22.09 & 16.76 & 277.4 \\
\cmidrule(l){4-8}
 &  &  & Generalized
  & Adjoint-FEM PWL & 19.61 & 14.88 & 463.2 \\
 &  &  &  & PINN PWL & 18.99 & 14.41 & 654.9 \\
 &  &  &  & PINN NN estimator & 19.06 & 14.46 & 356.4 \\
}
\begin{table}[H]
\centering
\caption{\RepresentativeTableCaption}
\label{\RepresentativeTableLabel}
\renewcommand{\arraystretch}{1.2}
\small
\setlength{\tabcolsep}{5pt}
\begin{tabular}{@{}c c c l l
                S[table-format=2.2]
                S[table-format=2.2]
                S[table-format=3.1]@{}}
\toprule
\textbf{Target} &
$\boldsymbol{\Delta t_{\mathrm{obs}}}$ &
\textbf{Noise} &
\textbf{Estimator} &
\textbf{Solver} &
{\shortstack{$\boldsymbol{\mathrm{RMSE}_{h}}$\\{\footnotesize(\si{\watt\per\meter\squared\per\kelvin})}}} &
{$\boldsymbol{\varepsilon_{h}}$ \textbf{(\%)}} &
{\textbf{Time (s)}} \\
\midrule
\RepresentativeTableRows
\bottomrule
\end{tabular}
\end{table}

At this Reynolds number, all three PINN reconstructions capture the
dominant decreasing trend of the reference HTC profile, although
noticeable differences occur in the downstream region. The generalized
PWL and NN-estimator reconstructions follow the reference profile more
closely in this region than the quadratic reconstruction.

The quantitative results show a more pronounced effect of the boundary
representation than observed for the $\mathrm{Re}=5000$ target. The
relative profile errors are \SI{16.76}{\percent}, \SI{14.41}{\percent}, and \SI{14.46}{\percent} for the
quadratic, PWL, and NN-estimator PINNs, respectively. The corresponding
Adjoint-FEM errors are \SI{18.96}{\percent} and \SI{14.88}{\percent} for the quadratic and PWL
representations. Thus, unlike the $\mathrm{Re}=5000$ case, the
generalized representations reduce the reconstruction error for this
target, with the PWL representation giving the largest reduction.

The computational cost shows a different pattern. The PWL PINN requires
the longest training time, whereas the NN-estimator achieves a nearly
identical reconstruction error with a shorter training time. For
Adjoint-FEM, the generalized PWL reconstruction also requires
substantially more computational time than the quadratic formulation.
The scratch-trained PWL and NN-estimator PINNs considered here are used
as the corresponding target models in the Gaussian-to-CHT
transfer-learning experiments discussed in
\Cref{subsec:transfer_profiles}.

\subsection{Transfer-learning performance}
\label{subsec:transfer_results}

Transfer performance is assessed by comparing each transferred target model
with a PINN trained from scratch using the same target observations and
boundary representation. The comparison considers both reconstruction
accuracy and target-training time, with the latter defined according to
\Cref{subsec:performance_metrics}.

\subsubsection{Transfer between noise levels at the same Reynolds number}
\label{subsec:transfer_noise}
This transfer setting examines whether a PINN trained on a noise-free source
problem can be reused when the measurement noise level changes. The source
and target correspond to the same CHT-derived physical case, with the target
observations generated by adding different levels of measurement noise.
The same $\mathrm{Re}=5000$ CHT profile is used throughout this
comparison, with the temporal observation interval fixed at
$\Delta t_{\mathrm{obs}}=\SI{1}{\second}$. The noise-free PWL and NN-estimator PINNs
serve as source models, and their transferred \SI{5}{\percent} and \SI{10}{\percent}-noise target
models are compared with independently trained PINNs using the same target observations and
boundary representations. The corresponding PWL Adjoint-FEM reconstructions are included as independent deterministic comparisons. Keeping the target profile and temporal observation interval fixed isolates
the comparison to measurement noise and the use of transfer learning.

\begin{figure}[!htbp]
\centering
\includegraphics{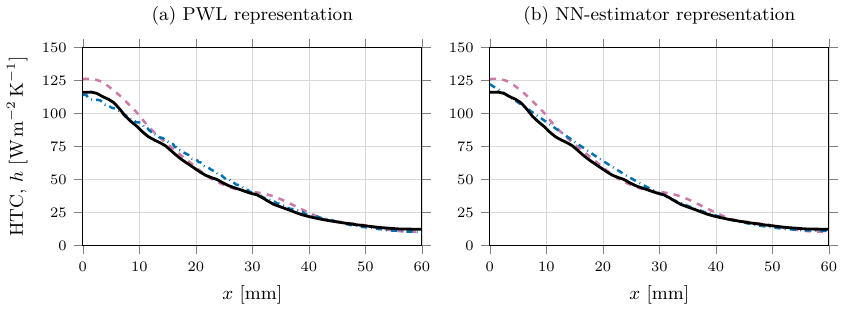}

\vspace{0.4em}
\includegraphics{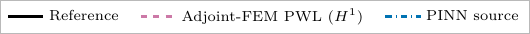}
\caption{Noise-free source-profile reconstruction for the
$\mathrm{Re}=5000$ case using the 1-s sampling interval: \textbf{(a)} PWL
and \textbf{(b)} NN-estimator source models (0\% rows of
\Cref{tab:transfer_noise_re5000}). The Adjoint-FEM PWL result is repeated in
both panels for comparison.}
\label{fig:fixed_Re_noise0_profiles}
\end{figure}

\begin{figure}[!hbtp]
\centering
\includegraphics{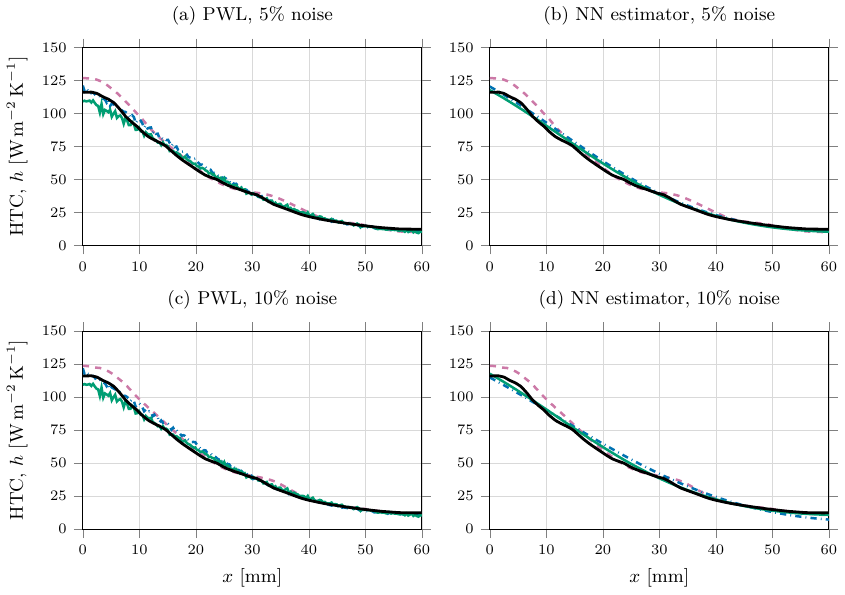}

\vspace{0.4em}
\includegraphics{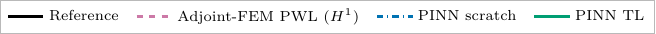}
\caption{Fixed-Reynolds-number transfer for the $\mathrm{Re}=5000$ case with
\textbf{(a,b)} 5\% and \textbf{(c,d)} 10\% measurement noise and the 1-s
sampling interval. Scratch and transfer-learned PINNs are compared within each
representation.}
\label{fig:fixed_Re_noisy_profiles}
\end{figure}

The noise-free source reconstructions are shown in
\Cref{fig:fixed_Re_noise0_profiles}. Both the PWL and NN-estimator PINNs
recover the principal spatial variation of the reference HTC profile,
providing the source models used for the subsequent transfer experiments.
The reconstructed profiles for the $5\%$ and $10\%$ measurement-noise cases
are shown separately in
\Cref{fig:fixed_Re_noisy_profiles}. The
transferred PWL and NN-estimator models remain close to their respective
scratch-trained reconstructions at both noise levels and reproduce the
dominant spatial variation of the reference HTC profile without
pronounced additional oscillations. The corresponding quantitative results are summarized in
\Cref{tab:transfer_noise_re5000}.

\def\RepresentativeTableCaption{Transfer across measurement-noise levels for
	the $\mathrm{Re}=5000$ CHT profile using the 1-s observation interval.}
\def\RepresentativeTableLabel{tab:transfer_noise_re5000}
\def\RepresentativeTableRows{
	$\mathrm{Re}=5000$ & 0\% & PWL
	& PINN & Source model & 3.51 & 6.02 & 207.0 & {---} \\
	&  &  & Adjoint-FEM & --- & 5.09 & 8.73 & 553.7 & {---} \\
	\cmidrule(l){3-9}
	&  & NN estimator
	& PINN & Source model & 3.18 & 5.46 & 333.8 & {---} \\
	\midrule
	$\mathrm{Re}=5000$ & 5\% & PWL
	& PINN & Scratch & 3.09 & 5.31 & 315.5 & {---} \\
	&  &  & PINN & TL from 0\% PWL source & 3.27 & 5.61 & 67.8 & 78.50 \\
	&  &  & Adjoint-FEM & --- & 5.10 & 8.75 & 537.6 & {---} \\
	\cmidrule(l){3-9}
	&  & NN estimator
	& PINN & Scratch & 3.02 & 5.19 & 356.4 & {---} \\
	&  &  & PINN & TL from 0\% NN source & 2.27 & 3.90 & 65.2 & 81.71 \\
	\midrule
	$\mathrm{Re}=5000$ & 10\% & PWL
	& PINN & Scratch & 3.19 & 5.47 & 329.5 & {---} \\
	&  &  & PINN & TL from 0\% PWL source & 3.20 & 5.49 & 69.4 & 78.95 \\
	&  &  & Adjoint-FEM & --- & 4.78 & 8.19 & 406.9 & {---} \\
	\cmidrule(l){3-9}
	&  & NN estimator
	& PINN & Scratch & 4.06 & 6.96 & 137.3 & {---} \\
	&  &  & PINN & TL from 0\% NN source & 2.25 & 3.85 & 57.9 & 57.86 \\
}
\begin{table}[!htbp]
	\centering
	\caption{\RepresentativeTableCaption}
	\label{\RepresentativeTableLabel}
	\renewcommand{\arraystretch}{1.2}
	\small
	\setlength{\tabcolsep}{2.4pt}
	\begin{tabular}{@{}c
			c
			l
			l
			>{\raggedright\arraybackslash}p{3.6cm}
			S[table-format=1.2]
			S[table-format=1.2]
			S[table-format=3.1]
			S[table-format=2.2]@{}}
		\toprule
		\textbf{Target} &
		\textbf{Noise} &
		\textbf{Estimator} &
		\textbf{Solver} &
		\textbf{Training case} &
		{\shortstack{$\boldsymbol{\mathrm{RMSE}_{h}}$\\{\footnotesize(\si{\watt\per\meter\squared\per\kelvin})}}} &
		{$\boldsymbol{\varepsilon_{h}}$ \textbf{(\%)}} &
		{\shortstack{\textbf{Time}\\\textbf{(s)}}} &
		{$\boldsymbol{S_t}$ \textbf{(\%)}} \\
		\midrule
		\RepresentativeTableRows
		\bottomrule
	\end{tabular}
\end{table}

Fixed-Reynolds-number transfer consistently reduces computational cost,
while its effect on reconstruction accuracy depends on the boundary
representation. At both noise levels, the transferred PWL models retain
reconstruction errors close to those obtained by scratch training, whereas
the transferred NN-estimator gives lower errors than scratch training in
the two cases considered. For the NN-estimator, the transferred
initialization therefore provides a useful starting point for target
adaptation, improving reconstruction accuracy in addition to reducing
training time. For example, at $\SI{5}{\percent}$ noise, NN
transfer reduces the production time from $\SI{356.4}{\second}$ to
$\SI{65.2}{\second}$ while reducing $\varepsilon_h$ from
$\SI{5.19}{\percent}$ to $\SI{3.90}{\percent}$. The PWL transfer shows a
similar computational reduction, with a $\SI{78.50}{\percent}$ saving at
the same noise level, while preserving comparable reconstruction accuracy.
Thus, for the same-Reynolds-number transfer pathway, reusing a model trained on
the same physical profile substantially reduces target training cost while the 
NN-estimator also benefits in reconstruction accuracy from the transferred initialization.

The profiles in \Cref{fig:fixed_Re_noisy_profiles} provide a visual
comparison of the transferred and scratch-trained reconstructions. Both
transferred representations recover the dominant spatial variation of the
reference HTC profile and remain visually close to their corresponding
scratch-trained reconstructions. Since only one noise realization is
considered at each noise level, these observations are specific to the
cases investigated here and do not constitute a statistical assessment of
noise robustness.

\subsubsection{Gaussian-to-CHT transfer across Reynolds numbers}
\label{subsec:transfer_profiles}
This transfer setting examines whether a PINN trained on the controlled
Gaussian source problem can be adapted to CFD--CHT-derived target problems
at different Reynolds numbers. Gaussian-trained models are transferred to
the \SI{5}{\percent}-noise CHT target cases at
$\mathrm{Re}=$\numlist{5000;10000;15000} using the
representation-specific procedures described in
\Cref{subsubsec:transfer_learning}. Each target is compared with a scratch-trained PINN using the same target observations and boundary representation. The
transferred models follow the specified transfer-learning configurations, whereas the scratch-trained models use the target-specific production
configurations reported in \Cref{tab:pinn_config}. The corresponding PWL Adjoint-FEM reconstructions
are included as independent deterministic comparisons.

\begin{figure}[!hbtp]
\centering
\includegraphics{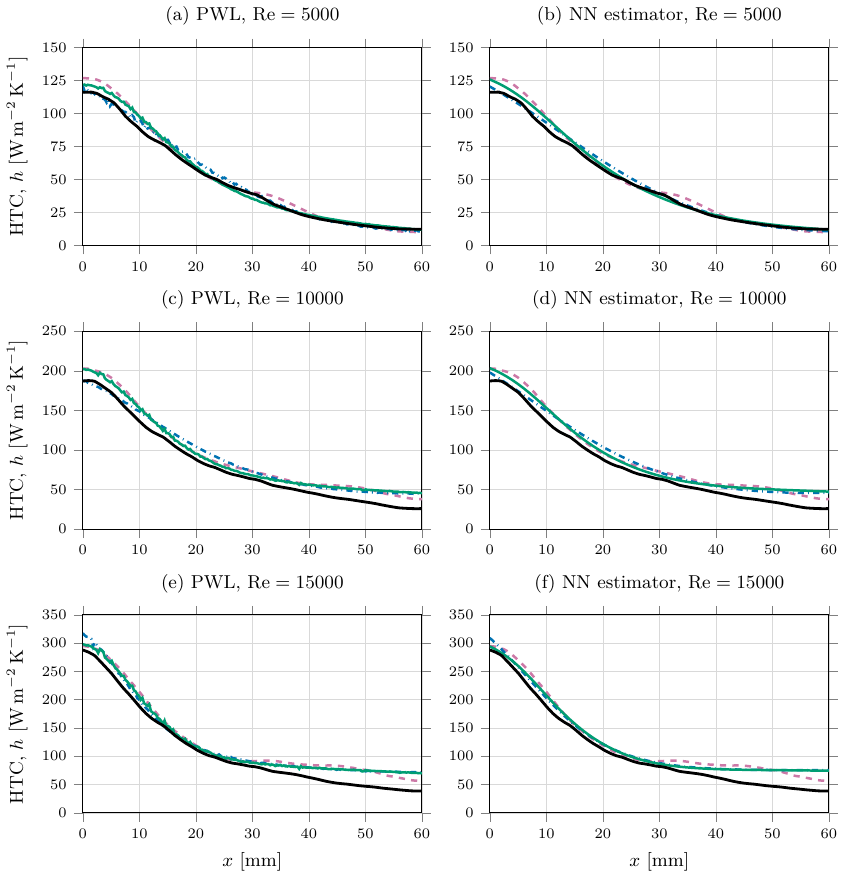}

\vspace{0.4em}
\includegraphics{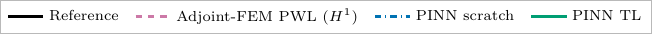}
\caption{Gaussian-to-CHT transfer at \textbf{(a,b)} $\mathrm{Re}=5000$,
\textbf{(c,d)} $\mathrm{Re}=10000$, and \textbf{(e,f)} $\mathrm{Re}=15000$
with 5\% measurement noise and the 1-s sampling interval. At $\mathrm{Re}=5000$, the transferred PWL curve is the 13,000-iteration
frozen-weight target initialized by the Gaussian parametric source.}
\label{fig:variable_Re_profiles}
\end{figure}

The reconstructed HTC profiles for the three Reynolds numbers are shown in
\Cref{fig:variable_Re_profiles}. The transferred and scratch-trained models
recover the dominant spatial variation of the reference HTC profiles. At
$\mathrm{Re}=10000$ and $\mathrm{Re}=15000$, noticeable discrepancies from
the reference profile remain in the downstream region for both transferred
and scratch-trained reconstructions. 

\def\RepresentativeTableCaption{Gaussian-to-CHT transfer at 5\% measurement
noise using the 1-s observation interval.}
\def\RepresentativeTableLabel{tab:transfer_profile_family}
\def\RepresentativeTableRows{
$\mathrm{Re}=5000$ & PWL
  & PINN & Scratch & 3.09 & 5.31 & 315.5 & {---} \\
&  & PINN & Frozen-weight TL from Gaussian parametric source & 3.98 & 6.83 & 86.2 & 72.67 \\
&  & Adjoint-FEM & --- & 5.10 & 8.75 & 537.6 & {---} \\
\cmidrule(l){2-8}
& NN estimator
  & PINN & Scratch & 3.02 & 5.19 & 356.4 & {---} \\
&  & PINN & TL from Gaussian NN source & 3.61 & 6.19 & 176.1 & 50.59 \\
\midrule
$\mathrm{Re}=10000$ & PWL
  & PINN & Scratch & 12.36 & 13.14 & 324.1 & {---} \\
&  & PINN & TL from Gaussian parametric source & 12.64 & 13.43 & 71.9 & 77.81 \\
&  & Adjoint-FEM & --- & 13.45 & 14.29 & 394.3 & {---} \\
\cmidrule(l){2-8}
& NN estimator
  & PINN & Scratch & 12.31 & 13.08 & 501.3 & {---} \\
&  & PINN & TL from Gaussian NN source & 12.80 & 13.60 & 84.8 & 83.09 \\
\midrule
$\mathrm{Re}=15000$ & PWL
  & PINN & Scratch & 18.99 & 14.41 & 654.9 & {---} \\
&  & PINN & TL from Gaussian parametric source & 18.62 & 14.13 & 98.7 & 84.94 \\
&  & Adjoint-FEM & --- & 19.61 & 14.88 & 463.2 & {---} \\
\cmidrule(l){2-8}
& NN estimator
  & PINN & Scratch & 19.06 & 14.46 & 356.4 & {---} \\
&  & PINN & TL from Gaussian NN source & 18.93 & 14.36 & 73.1 & 79.49 \\
}
\begin{table}[!hbtp]
\centering
\caption{\RepresentativeTableCaption}
\label{\RepresentativeTableLabel}
\renewcommand{\arraystretch}{1.2}
\small
\setlength{\tabcolsep}{3.4pt}
\begin{tabular}{@{}c
                l
                l
                >{\raggedright\arraybackslash}p{3.5cm}
                S[table-format=2.2]
                S[table-format=2.2]
                S[table-format=3.1]
                S[table-format=2.2]@{}}
\toprule
\textbf{Target} &
\textbf{Estimator} &
\textbf{Solver} &
\textbf{Training case} &
{\shortstack{$\boldsymbol{\mathrm{RMSE}_{h}}$\\{\footnotesize(\si{\watt\per\meter\squared\per\kelvin})}}} &
{$\boldsymbol{\varepsilon_{h}}$ \textbf{(\%)}} &
{\shortstack{\textbf{Time}\\\textbf{(s)}}} &
{$\boldsymbol{S_t}$ \textbf{(\%)}} \\
\midrule
\RepresentativeTableRows
\bottomrule
\end{tabular}
\end{table}

Gaussian-to-CHT transfer exhibits a stronger dependence on the target. At
$\mathrm{Re}=5000$, frozen-weight transfer from the Gaussian parametric source
to the PWL target increases $\varepsilon_h$ from 5.31 to 6.83\%, while reducing
the target time from 315.5 to 86.2~s (72.67\%). The Gaussian NN-source transfer
increases the error from 5.19 to 6.19\% and saves 50.59\%. This contrasts with the same-Reynolds-number NN transfer, where transfer reduced both target-training time and reconstruction error. Thus, when the source and target inverse problems differ, transfer can substantially reduce training time while producing either comparable or different reconstruction accuracy relative to scratch training.

The higher-Re targets show a more favorable speed--accuracy balance. At
$\mathrm{Re}=10000$, the transferred PWL and NN-estimator errors differ from
their scratch counterparts by $+0.29$ and $+0.52$ percentage points, while
saving 77.81 and 83.09\% of the training time. At $\mathrm{Re}=15000$, the
differences are $-0.28$ and $-0.10$ percentage points, with savings of 84.94
and 79.49\%. Thus, for these targets, Gaussian-source transfer largely
preserves the accuracy attainable by scratch training while reducing
target-estimation time by approximately 78--85\%. The corresponding Adjoint-FEM results provide an independent deterministic comparison for these transferred reconstructions: their relative
errors are 6.19--6.83\%, 13.43--13.60\%, and 14.13--14.36\% at increasing
Reynolds number, compared with 8.75, 14.29, and 14.88\% for Adjoint-FEM.

The Gaussian-to-CHT results therefore show that transfer substantially reduces target-training cost, while the effect on reconstruction accuracy depends on the target problem. At $\mathrm{Re}=5000$, transfer produces a larger accuracy penalty, whereas the $\mathrm{Re}=10000$ and $\mathrm{Re}=15000$ transferred models remain close to their scratch-trained counterparts. Because the source and target problems differ in both boundary-profile characteristics and data-generation setting, these results demonstrate computational reuse across different profile families without implying identical reconstruction accuracy.

The reported savings therefore represent reductions in target-training time
under the timing convention defined in \Cref{subsec:performance_metrics}.
They do not represent reductions in the total computational cost of
developing and applying the transfer-learning framework, since source-model
training and other offline costs are excluded from the reported times.
For repeated inverse estimation, these offline costs can be amortized over
multiple target reconstructions. For a single isolated inverse solve,
however, they would need to be included in a complete computational-cost
assessment.

%

\section{Conclusions}
\label{sec:conclusions}
This study investigated generalized boundary representations and transfer
learning for PINN-based estimation of spatially varying convective heat
transfer coefficients in jet impingement cooling. Building on the
inverse heat-transfer framework established in our previous work~\citep{hazra_pinnbased_2026a}, the study
extended the previously used quadratic HTC representation to piecewise-linear
(PWL) and neural-network boundary estimators and examined transfer learning
under different source--target conditions. In addition, a discrete Adjoint-FEM formulation with \(H^1\)-seminorm regularization is employed to provide an independent deterministic comparison of reconstruction accuracy and computational cost.

For the representative \(\mathrm{Re}=5000\) target, the generalized PWL and NN-estimator PINNs reproduced the dominant spatial decay of the reference HTC profile with reconstruction errors comparable to that of the quadratic representation. Increasing the sampling interval from \(\SI{1}{\second}\) to \(\SI{4}{\second}\) resulted in representation-dependent changes in reconstruction accuracy, with no consistent improvement or deterioration across the three representations. For the \(\mathrm{Re}=15000\) target, the generalized representations yielded lower reconstruction errors than the quadratic representation, with the PWL and NN-estimator PINNs achieving errors of \(14.41\%\) and \(14.46\%\), respectively, compared with \(16.76\%\) for the quadratic representation. The benefit of generalized boundary representations was therefore not uniform across the considered targets: the representations produced comparable errors at \(Re=5000\), whereas the higher-Reynolds-number target showed a larger reduction in error with the generalized representations.

Transfer learning substantially reduced target-training time in both the
fixed-Reynolds-number and Gaussian-to-CHT transfer pathways. For the
fixed-Reynolds-number $\mathrm{Re}=5000$ case, transfer from the noise-free
source reduced target-training time by
$\SIrange{57.86}{81.71}{\percent}$ across the PWL and NN-estimator
representations and the two noise levels considered. The transferred PWL
models retained reconstruction errors within $0.30$ percentage points of
their scratch-trained counterparts, while the NN-estimator transfers also
reduced reconstruction error for the two noise levels considered.
Gaussian-to-CHT transfer similarly reduced target-training time, with
savings of $\SIrange{50.59}{84.94}{\percent}$ across the Reynolds numbers
and representations considered. Its effect on reconstruction accuracy was
target dependent: at $\mathrm{Re}=5000$, the transferred PWL and
NN-estimator models showed increases of $1.52$ and $1.00$ percentage
points, respectively, relative to scratch training, whereas the
differences at $\mathrm{Re}=10000$ and $\mathrm{Re}=15000$ remained within
$0.52$ percentage points. Thus, the computational benefit of transfer was
more consistent than its effect on reconstruction accuracy.

The principal result is that flexibility and computational reuse address
different limitations of PINN-based inverse estimation: generalized
representations reduce dependence on an imposed boundary-profile family,
whereas transfer learning reduces the training cost of solving related
target inverse problems. The Gaussian-to-CHT results further show that
substantial target-training savings can be obtained even when the source
and target differ in both boundary-profile characteristics and
data-generation setting.

The present study provides a basis for several extensions that are being
addressed in ongoing and future work. These include investigating the effects
of sensor placement and PWL spatial resolution, extending transfer learning
to a broader range of source--target conditions, and assessing the framework
using repeated noise realizations and experimental measurements. 
The reduced target-training time achieved through transfer learning also provides a
basis for real-time thermal monitoring and digital-twin applications, where
repeated inverse estimation is required. Future work will extend the inverse formulation to experimental data and more general heat-transfer coefficients with spatio-temporal and temperature dependence.

\section*{Acknowledgements}
Arijit Hazra would like to acknowledge funding support from the Ramanujan Fellowship (RJF/2022/000046) administered by Anusandhan National Research Foundation (ANRF), India.

\appendix

\appendix
\section{Adjoint-FEM Regularization and Optimization}
\label{app:adjoint_fem}

This appendix provides the numerical details used for the generalized
piecewise-linear (PWL) Adjoint-FEM reconstructions, including the
parameter bounds, regularization selection, and optimization stopping
criterion. The discrete adjoint formulation itself is given in
\Cref{subsec:adjoint}.

\subsection{PWL parameterization and regularization}
\label{subsec:app_pwl_regularization}
The generalized boundary profile is represented using a piecewise-linear
parameterization,
\begin{align}
	\Bi_{fc}(x)
	=
	\sum_{j=0}^{N_p-1} b_j\phi_j(x),
	\label{eq:app_pwl}
\end{align}
where $b_j$ are the nodal Biot-number coefficients and $\phi_j(x)$ are
the corresponding piecewise-linear basis functions.
The boundary coefficients are constrained according to
\begin{align}
	0 \leq b_j \leq 2,
	\qquad j=0,\ldots,N_p-1,
	\label{eq:app_pwl_bounds}
\end{align}
and the initial boundary profile is prescribed as
\begin{align}
	\Bi_{fc}^{(0)}(x)=1-x.
	\label{eq:app_initial_profile}
\end{align}
The coefficient bounds are imposed directly by the L-BFGS-B
optimization.

For the generalized PWL reconstructions, first-order Tikhonov
regularization based on the $H^1$ seminorm is introduced to penalize
rapid spatial variation of the reconstructed boundary profile,
\begin{align}
	J_{\mathrm{reg}}
	=
	\frac{\mu_{\mathrm{reg}}}{2}
	\int_{\Gamma_c}
	\left|
	\frac{\partial \Bi_{fc}}{\partial x}
	\right|^2
	\,\mathrm{d}\Gamma,
	\label{eq:app_h1_regularization}
\end{align}
where $\mu_{\mathrm{reg}}$ is the regularization parameter. The
regularized inverse objective is therefore
\begin{align}
	J(\mathbf b)
	=
	J_{\mathrm{data}}(\mathbf b)
	+
	J_{\mathrm{reg}}(\mathbf b).
	\label{eq:app_regularized_objective}
\end{align}
The regularization penalizes the spatial gradient of the boundary
profile rather than its magnitude. No zeroth-order $L^2$ penalty or
reference-profile prior is included.

Using the PWL representation in \Cref{eq:app_pwl}, the contribution of
the regularization term to the gradient with respect to an individual
boundary coefficient is
\begin{align}
	\frac{\partial J_{\mathrm{reg}}}{\partial b_j}
	=
	\mu_{\mathrm{reg}}
	\int_{\Gamma_c}
	\frac{\partial \Bi_{fc}}{\partial x}
	\frac{\partial \phi_j}{\partial x}
	\,\mathrm{d}\Gamma.
	\label{eq:app_reg_gradient}
\end{align}
The complete gradient supplied to the L-BFGS-B optimizer consists of
the discrete-adjoint contribution from the data-misfit term and the
regularization contribution in \Cref{eq:app_reg_gradient}.

The regularization parameter is examined using L-curve diagnostics to
assess the trade-off between data fit and profile regularity
\citep{hansen_oleary_lcurve_1993}. Following the practice of logarithmic
sampling of regularization parameters illustrated in
\citet{hansen_regularization_tools_2008}, a common seven-point grid is
defined by
\begin{align}
	\mu_{\mathrm{reg},j}=10^{-3+5j/11},
	\qquad j=0,\ldots,6.
	\label{eq:app_mu_grid}
\end{align}
The sampled values are listed in \Cref{eq:seven_point_grid}.
All seven evaluated samples are included in the L-curve plots, which
show the trade-off between the temperature data-misfit and the spatial
regularity of the reconstructed boundary profile.

For the $Re=5000$ cases with \SI{1}{\second} sampling intervals,
the resulting L-curves for \SI{0}{\percent}, \SI{5}{\percent}, and
\SI{10}{\percent} measurement noise are shown in
\Cref{fig:appendix_lcurve_noise}. The corresponding L-curves for the remaining
\SI{5}{\percent}-noise cases are shown in
\Cref{fig:appendix_lcurve_other_cases}. The sampled value
\begin{align}
	\mu_{\mathrm{reg}}=0.02310130
	\label{eq:app_mu_selected}
\end{align}
is adopted for all reported generalized PWL reconstructions.

\newcommand{\AppendixLCurvePanel}[1]{%
  \begin{minipage}[t]{0.31\textwidth}
    \centering
    \includegraphics{#1}
  \end{minipage}%
}

\begin{figure}[H]
  \centering
  \AppendixLCurvePanel{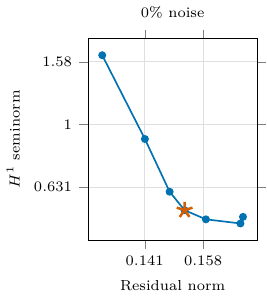}\hfill
  \AppendixLCurvePanel{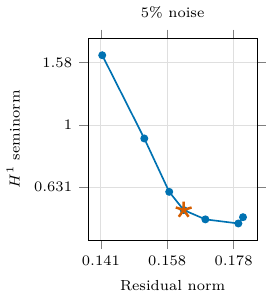}\hfill
  \AppendixLCurvePanel{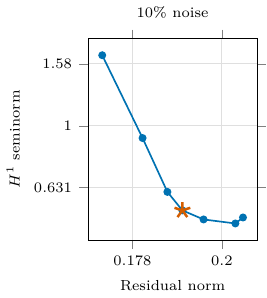}
  \caption{$H^1$-seminorm Adjoint-FEM PWL L-curves for the
  $\mathrm{Re}=5000$ cases with the 1-s sampling interval. Every sweep starts from
  $\Bi_{fc,0}(x)=1-x$ and uses the fixed bounds $0\leq\Bi_{fc}\leq2$.
  The red star marks the value
  $\mu_{\mathrm{reg}}=2.31013\times10^{-2}$ adopted for all production
  reconstructions.}
  \label{fig:appendix_lcurve_noise}
\end{figure}

\begin{figure}[H]
  \centering
  \AppendixLCurvePanel{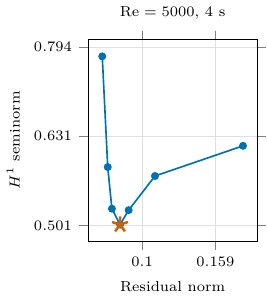}\hfill
  \AppendixLCurvePanel{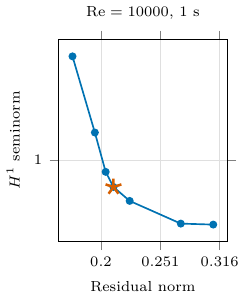}\hfill
  \AppendixLCurvePanel{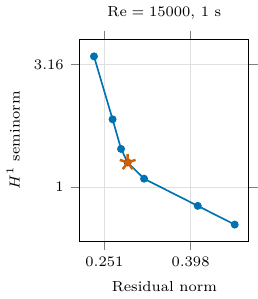}
  \caption{$H^1$-seminorm Adjoint-FEM PWL L-curves for the remaining 5\%-noise
  cases. The initialization, bounds, candidate grid, and marked production
  value are identical to \Cref{fig:appendix_lcurve_noise}.}
  \label{fig:appendix_lcurve_other_cases}
\end{figure}

\subsection{Optimization and stopping criterion}
\label{subsec:app_optimization}

The production PWL inverse problems are solved using the bound-constrained
L-BFGS-B algorithm. The stopping criterion is applied to accepted
optimization iterates rather than to every objective evaluation
performed by the optimizer. Let $J_k$ and $\mathbf b_k$ denote the
objective value and the $N_p$-component boundary-control vector at the
$k$th accepted iterate. The relative change in the objective over a
window of $w$ accepted iterations is defined as
\begin{align}
	r_J^{(k)}
	=
	\frac{
		\left|J_k-J_{k-w}\right|
	}{
		\max\left(\left|J_{k-w}\right|,\epsilon\right)
	},
	\label{eq:app_objective_stop}
\end{align}
while the corresponding change in the reconstructed boundary profile is
measured using
\begin{align}
	r_b^{(k)}
	=
	\frac{
		\left\|\mathbf b_k-\mathbf b_{k-w}\right\|_W
	}{
		\max\left(
		\left\|\mathbf b_{k-w}\right\|_W,\epsilon
		\right)
	},
	\label{eq:app_profile_stop}
\end{align}
where $\epsilon$ is a numerical safeguard and
$\|\cdot\|_W$ denotes the spatially weighted norm
\begin{align}
	\|\mathbf v\|_W
	=
	\left(
	\sum_i \omega_i v_i^2
	\right)^{1/2},
	\label{eq:app_weighted_norm}
\end{align}
with $\omega_i$ denoting the composite-trapezoidal weights associated
with the PWL boundary nodes.

The stopping window comprises three accepted iterations ($w=3$).
A convergence check is performed only after at least ten accepted
optimizer iterations. The check is satisfied when
\begin{align}
	r_J^{(k)} < 5\times10^{-3},
	\qquad
	r_b^{(k)} < 5\times10^{-3}.
	\label{eq:app_stop_criterion}
\end{align}
Termination occurs when this joint criterion is satisfied for two
consecutive checks. Thus, both the inverse objective and the reconstructed
boundary profile must stabilize before the optimization is terminated.

The resulting terminal accepted iterations for the six production cases
are summarized in \Cref{tab:app_adjoint_iterations}. Five cases satisfy
the adaptive stopping criterion. The $Re=5000$ case with a
\SI{4}{\second} sampling interval reaches the prescribed maximum of
50 accepted iterations after satisfying the convergence criterion once
and is therefore classified as maximum-iteration termination.

\begin{table}[htbp]
	\centering
	\caption{%
		Termination of the generalized PWL Adjoint-FEM
		optimizations.%
	}
	\label{tab:app_adjoint_iterations}
	\begin{tabular}{cccc}
		\hline
		$Re$ &
		Sampling interval &
		Noise level &
		Accepted iterations \\
		\hline
		5000  & \SI{1}{\second} & \SI{0}{\percent}  & 41 \\
		5000  & \SI{1}{\second} & \SI{5}{\percent}  & 43 \\
		5000  & \SI{1}{\second} & \SI{10}{\percent} & 33 \\
		5000  & \SI{4}{\second} & \SI{5}{\percent}  & 50$^{a}$ \\
		10000 & \SI{1}{\second} & \SI{5}{\percent}  & 33 \\
		15000 & \SI{1}{\second} & \SI{5}{\percent}  & 39 \\
		\hline
	\end{tabular}

	\vspace{1mm}
	\raggedright
	\footnotesize
	$^{a}$Maximum iteration limit reached after one successful
	stopping check.
\end{table}

\begin{figure}[!hbtp]
	\centering
	\includegraphics{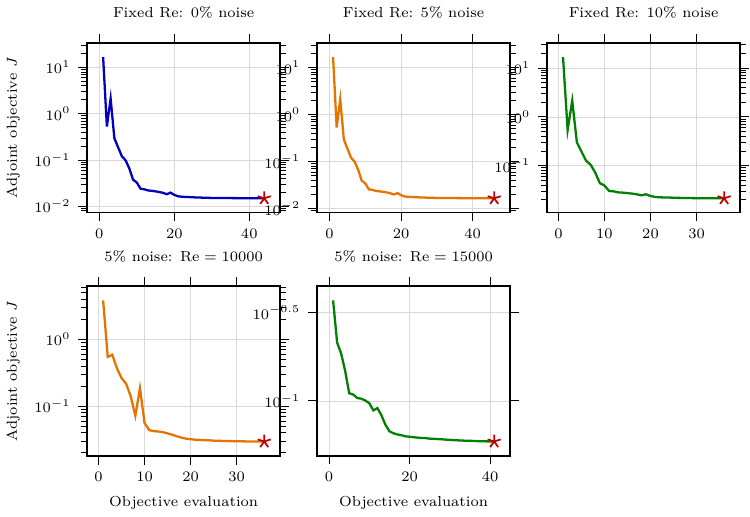}
	\caption{Adjoint-FEM PWL objective histories for the comparisons in~\Cref{tab:transfer_noise_re5000,tab:transfer_profile_family}. The five distinct cases with the 1-s sampling interval are shown; the shared $\mathrm{Re}=5000$, \SI{5}{\percent}-noise case appears once, and the \SI{4}{\second} case is omitted. Objectives are plotted on a logarithmic scale, with red stars marking the final stored evaluations.}
	\label{fig:appendix_adjoint_loss}
\end{figure}

The objective histories in \Cref{fig:appendix_adjoint_loss} retain all
stored L-BFGS-B evaluations without resampling. Each history includes two
preliminary evaluations that are not counted as accepted optimization
iterations; the final evaluation number therefore exceeds the
accepted-iteration count in \Cref{tab:app_adjoint_iterations} by two.

\section{PINN Optimization and Convergence}
\label{app:pinn_diagnostics}

This appendix provides additional details on the PINN stopping criterion and
the convergence diagnostics used in the reported calculations. All PINN configurations used in the reported comparisons employ early
stopping after a minimum of \num{5000} training iterations. Because the
inverse problem is concerned with reconstruction of the unknown boundary
coefficient, convergence is monitored through stabilization of the
estimated boundary parameter rather than solely through stagnation of the
total PINN loss. For coefficient-based estimators, the monitored quantity
is the vector of estimated boundary coefficients or reconstructed-profile
values. For the NN estimator, the monitored quantity is the spatial
integral of the sampled boundary profile, evaluated using composite
Simpson quadrature. Denoting the monitored quantity by $\mathbf{s}_k$,
with a single component in the NN-estimator case, a convergence check at
iteration $k$ is satisfied when
\begin{align}
	\left|s_{k,i}-s_{k-\Delta k,i}\right|
	\leq
	\tau_{\mathrm{abs}}
	+
	\tau_{\mathrm{rel}}
	\max\left(
	|s_{k,i}|,
	|s_{k-\Delta k,i}|,
	1
	\right),
	\qquad \forall i .
	\label{eq:appendix_pinn_early_stop}
\end{align}

The stability counter is increased by $\Delta k$ when
\Cref{eq:appendix_pinn_early_stop} is satisfied and reset to zero when
the condition is violated. Training terminates when the counter reaches
the prescribed patience $P$. Across the production configurations, the
stopping parameters were selected from
\begin{align}
	P
	&\in
	\{500,1000\},
	&
	\tau_{\mathrm{abs}}
	&\in
	\left\{
	\SI{1e-5}{},
	\SI{5e-5}{},
	\SI{5e-4}{}
	\right\},
	&
	\tau_{\mathrm{rel}}
	&\in
	\left\{
	\SI{1e-6}{},
	\SI{1e-5}{},
	\SI{1e-4}{}
	\right\}.
	\label{eq:appendix_pinn_stop_values}
\end{align}

The stopping parameters were specified for each production training
configuration to accommodate differences in convergence behavior.

\begin{figure}[!hbtp]
  \centering
  \includegraphics{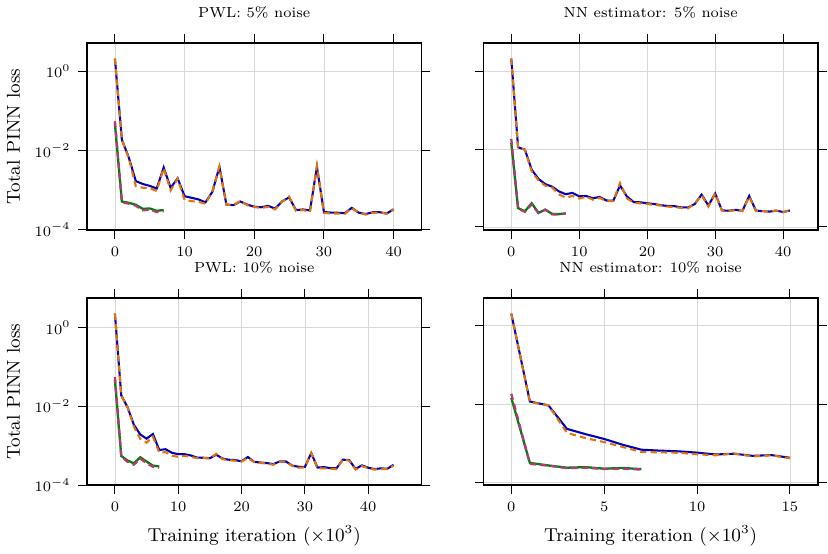}

  \vspace{0.25em}
  {\footnotesize
  \begin{tabular}{@{}ll@{\qquad}ll@{}}
    \SelectedLegendLine{blue!75!black,line width=1pt,solid} & Scratch train &
    \SelectedLegendLine{orange!90!black,line width=1pt,densely dashed} & Scratch test \\
    \SelectedLegendLine{green!50!black,line width=1.1pt,solid} & Transfer train &
    \SelectedLegendLine{magenta!70!black,line width=1.1pt,densely dashed} & Transfer test
  \end{tabular}}
  \caption{Complete PINN train/test histories for the fixed-Reynolds-number transfer
  targets in \Cref{tab:transfer_noise_re5000}. Each ordinate is the sum
  of the PDE, four boundary-condition, initial-condition, and observation
  losses shared by both boundary representations. Curves retain every stored
  1000-iteration checkpoint and terminate at the model-configured stopping
  iteration.}
  \label{fig:appendix_table8_pinn_loss}
\end{figure}

\begin{figure}[!hbtp]
  \centering
  \includegraphics{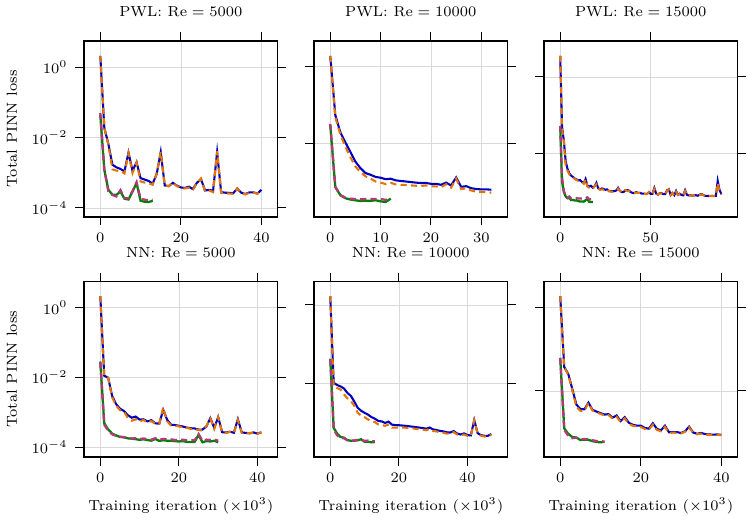}

  \vspace{0.25em}
  {\footnotesize
  \begin{tabular}{@{}ll@{\qquad}ll@{}}
    \SelectedLegendLine{blue!75!black,line width=1pt,solid} & Scratch train &
    \SelectedLegendLine{orange!90!black,line width=1pt,densely dashed} & Scratch test \\
    \SelectedLegendLine{green!50!black,line width=1.1pt,solid} & Gaussian-source transfer train &
    \SelectedLegendLine{magenta!70!black,line width=1.1pt,densely dashed} & Gaussian-source transfer test
  \end{tabular}}
  \caption{Complete PINN train/test histories for all Gaussian-to-CHT transfer
  targets in \Cref{tab:transfer_profile_family}. The total PINN loss uses the
  same seven components as \Cref{fig:appendix_table8_pinn_loss}. The
  $\mathrm{Re}=5000$ PWL transfer is the frozen-weight transfer from the
  Gaussian parametric source model reported in \Cref{tab:transfer_profile_family}.}
  \label{fig:appendix_table10_pinn_loss}
\end{figure}

Training and test losses were recorded every \num{1000} iterations
for the scratch-trained and transferred target runs. The total PINN loss shown in
\Cref{fig:appendix_table8_pinn_loss,fig:appendix_table10_pinn_loss}
is the sum of the PDE residual, four boundary-condition,
initial-condition, and observation terms. These seven components are
present in the PWL and NN-estimator configurations; representation-specific
regularization terms are excluded from these plots so that the definition
of the displayed total loss remains consistent across representations.
The loss histories are used as within-run convergence diagnostics rather
than as cross-run measures of reconstruction accuracy, which is assessed
using the HTC-profile errors and observation-data discrepancies reported
in the main results.

\clearpage



\end{document}